\documentclass[10pt,twocolumn,letterpaper]{article}

\usepackage{multirow}
\usepackage{overpic}
\usepackage[pagenumbers]{iccv} 
\newcommand{\name}{\emph{Metric-Solver}}

\def\neardepth{scaled near depth\xspace} %
\def\fardepth{tapered far depth\xspace} %
\definecolor{iccvblue}{rgb}{0.21,0.49,0.74}
\usepackage[pagebackref,breaklinks,colorlinks,allcolors=iccvblue]{hyperref}

\title{\name : Sliding Anchored Metric Depth Estimation from a Single Image}

\author{{Tao Wen$^{1,\ast}$, 
Jiepeng Wang$^{2,\ast,\ddagger}$, Yabo Chen$^{3}$, Shugong Xu$^{4,\dagger}$, Chi Zhang$^{2,\dagger}$, Xuelong Li$^{2,\dagger}$ } \\
$^1$Shanghai University, $^2$Institute of Artificial Intelligence, China Telecom (TeleAI),  \\ $^3$Shanghai Jiaotong University,
$^4$Xi'an Jiaotong-Liverpool University
}

\begin{document}

\vspace{-8pt}
\twocolumn[{%
        \vspace{-25pt}
	\maketitle
	\renewcommand\twocolumn[1][]{#1}%
	\begin{center}
		\centering
        \includegraphics[width=1.0\linewidth]{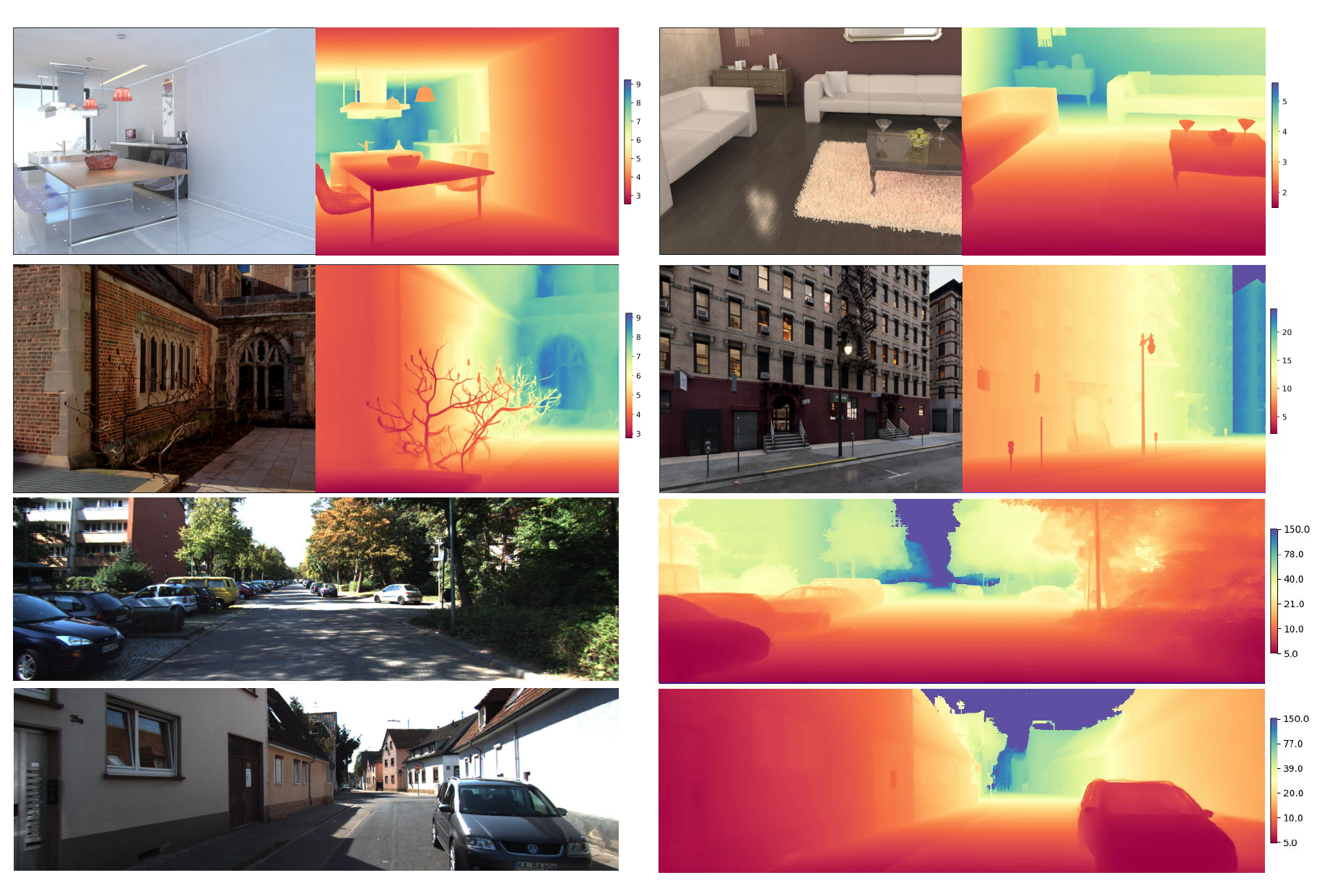}
        \vspace{-15pt}
        \captionof{figure}{\textbf{A gallery of our predictions across various scenarios.} The Metric-Solver model effectively addresses different in-the-wild scenes with unknown camera settings. This model delivers precise metric depth predictions across a variety of scenarios, including but not limited to indoor and outdoor scenes, autonomous driving scenarios, and various datasets which are captured by different cameras. The side bar along each depth map indicates the predicted depth range in meters.
         } 
		\label{fig:teaser}
	\end{center}
}
]

\maketitle

\renewcommand{\thefootnote}{\fnsymbol{footnote}}
\footnotetext[1]{Equal contribution, $^\dagger$Corresponding author, $^\ddagger$Project lead.}

\begin{abstract}

Accurate and generalizable metric depth estimation is crucial for various computer vision applications but remains challenging due to the diverse depth scales encountered in indoor and outdoor environments. In this paper, we introduce Metric-Solver, a novel sliding anchor-based metric depth estimation method that dynamically adapts to varying scene scales.
Our approach leverages an anchor-based representation, where a reference depth serves as an anchor to separate and normalize the scene depth into two components: scaled near-field depth and tapered far-field depth. The anchor acts as a normalization factor, enabling the near-field depth to be normalized within a consistent range while mapping far-field depth smoothly toward zero. Through this approach, any depth from zero to infinity in the scene can be represented within a unified representation, effectively eliminating the need to manually account for scene scale variations.
More importantly, for the same scene, the anchor can slide along the depth axis, dynamically adjusting to different depth scales. A smaller anchor provides higher resolution in the near-field, improving depth precision for closer objects while a larger anchor improves depth estimation in far regions. 
This adaptability enables the model to handle depth predictions at varying distances and ensure strong generalization across datasets. 
Our design enables a unified and adaptive depth representation across diverse environments. Extensive experiments demonstrate that Metric-Solver outperforms existing methods in both accuracy and cross-dataset generalization. Our project page: \url{https://tele-ai.github.io/MetricSolver/}

\end{abstract}    
\section{Introduction}
\label{sec:intro}

Monocular depth estimation from a single image~\cite{eigen2014depth, fu2018deep, adabins,newcrfs,zerodepth,zoedepth,ait,iebins,depth_anything}, plays a crucial role in a wide range of computer vision applications, including robotics~\cite{pddepth}, augmented reality~\cite{real}, 3D graphics~\cite{3dgs,nerf} and autonomous driving~\cite{autotrack}. 
Depth estimation methods can be broadly divided into two types: relative depth estimation~\cite{midas,depth_anything} and metric depth estimation~\cite{metric3d,metric3dv2,zerodepth,zoedepth}. 
Relative depth estimation methods predict the depth of objects in a scene relative to one another, providing spatial relationships between objects. 
In contrast, metric depth estimation aims to predict the true, real-world scale of the scene, providing accurate measurements of the distance between objects and the camera. 
Despite its advantages, metric depth estimation presents significant challenges, particularly in terms of scale variation across different datasets~\cite{zoedepth,adabins,newcrfs}, which cause metric ambiguity due to mixed-data training. 
Recently, many methods have been proposed to address the generalization problem in metric depth estimation by using pre-input camera intrinsics to simplify the issue, including Metric3D~\cite{metric3dv2} and Depth Any Camera~\cite{depth_any_camera}, which have achieved significant performance improvements. However, for the aforementioned methods, in-the-wild images with unknown camera settings remain an challenging problem.

Another key difficulties in metric depth estimation is handling the varying depth scales across different in-the-wild scenes, such as indoor and outdoor environments. In indoor scenes, the maximum depth is typically within several meters, while in outdoor scenes, it can extend to several hundred meters. This disparity makes it challenging to use a unified normalization approach across diverse scenes, leading to issues in network training and generalization. 

To address these challenges, we propose Metric-Solver, a novel sliding anchor-based metric depth representation that dynamically adjusts to varying depth scales. To represent depth values ranging from zero to infinity in a unified manner, we introduce a reference depth as an anchor, which partitions the scene depth into two distinct components: scaled near-field depth and tapered far-field depth, with the anchor depth acting as the normalization factor.
Specifically, for depth values within the anchor range, we apply linear normalization to obtain the scaled near-field depth, ensuring depth values remain within a consistent range. For depth values beyond the anchor, we use an exponential normalization function to smoothly compress far-depth values, where the anchor depth is mapped to 1 and infinity gradually decays to 0. 
This transformation preserves depth continuity while allowing the model to handle far-field depth variations effectively.
Besides, by dynamically sliding the anchor along the depth axis, our method allows the model to adaptively adjust depth scale within the same scene. For instance, a smaller anchor enhances near-field depth fidelity, improving the accuracy of predictions for closer objects, while a larger anchor better captures far-field depth relationships.
This adaptability not only enhances the model’s precision in different depth regions but also improves its versatility across various scene scales.

To implement the sliding anchor-based metric depth estimation, we adopt a \emph{one-encoder, two-decoder} architecture, where the latent features extracted by the encoder are processed by two separate decoder branches to predict \neardepth and \fardepth.
Specifically, we begin by utilizing a pretrained image encoder~\cite{dinov2} to extract latent features from the input image. These features, along with the anchor depth information, are then fed into two independent decoding branches: one responsible for predicting \neardepth, which captures depth values within the anchor range, and the other for predicting \fardepth, which models depth values beyond the anchor using an exponential transformation.
Finally, the outputs from both branches are re-projected into real-world depth values using the reference anchor depth, fusing into a complete metric depth estimation. During training, the reference anchor depth is randomly selected for each input image, allowing the model to adapt to varying scene scales and enhancing its ability to generalize across different environments.

We validate our method on various benchmark datasets, including both indoor~\cite{nyud,ibims,hypersim,sunrgbd} and outdoor scenes~\cite{kitti,vkitti,diode,synthia}, to assess its robustness and generalization capability. Extensive experiments demonstrate the model's strong performance across diverse datasets and its ability to handle zero-shot settings. 
Our results emphasize the adaptability of the sliding anchor-based depth estimation approach, which consistently delivers accurate metric depth predictions across a range of scene scales.

To summarize, we present the following key contributions:
\begin{itemize}

    \item We proposed a novel framework, i.e. the sliding anchored method, which effectively models metric ambiguity in in-the-wild images with unknown camera settings.

    \item We designed a one-encoder, two-decoder architecture that effectively bridges the varying depth scales across indoor and outdoor scenes.

    \item Our method demonstrates strong generalization ability across various benchmark datasets, achieving state-of-the-art (SOTA) performance on all seven benchmarks.

\end{itemize}

\section{Related Works}
\label{sec:related_works}

Recovering the depth of each pixel in the scene from a single camera view is one of the most fundamental topics in vision tasks~\cite{eigen2014depth}.  In this section, we review the recent process of this research areas, including relative depth estimation and metric depth estimation.

\subsection{Relative Depth Estimation}

Relative depth estimation focuses on predicting the depth relationships between objects in a scene rather than their absolute distances. This approach is widely used to handle the high dynamic range of depth distributions in indoor and outdoor environments, where depth values can vary significantly due to differences in scene structure, lighting, and camera intrinsic parameters~\cite{newcrfs}. To address these challenges, many methods~\cite{monocular, midas, midasv31, dpt, SIDEreviewMERTAN2022103441} normalize depth values and use them as learning targets for neural networks.
In traditional depth estimation algorithms~\cite{saxena2005learning,nagai2002hmm,michels2005high}, they typically rely on dense depth regression using hand-crafted features. However, due to the limited expressiveness of these features, such methods struggle in complex environments and low-texture regions, resulting in poor depth estimation performance.

Deep-learning-based depth estimation approaches can be broadly categorized into discriminative models that employ depth regression decoders~\cite{zerodepth,zoedepth,dpt,midas,ait,p3depth,depth_anything} and generation models\cite{marigold, depthfm, geowizard}, which leverage diffusion models\cite{sd} for depth prediction.
In discriminative models, powerful backbone networks~\cite{dinov2} extract multi-scale features from the image, and the decoder decodes the depth information. Eigen et al~\cite{eigen2014depth} used scale-invariant loss to train relative depth estimation networks, allowing the network to focus on predicting the relative order of pixels' depths rather than being constrained by the absolute scale of the scene. Some works improve depth estimation accuracy by introducing additional priors such as segmentation maps~\cite{zhang2018joint}. To address scene generalization and edge blurring issues, the Depth Anything series~\cite{depthanythingv1,depthanythingv2} use semi-supervised strategies to complete multi-scene learning on large datasets, significantly enhancing the network's ability to refine depth at scene edges by incorporating synthetic datasets~\cite{hypersim,vkitti}. 
In generative models, Marigold~\cite{marigold} encodes RGB images and depth separately into latent space, uses the latent code of RGB as a condition, and denoises the noisy latent code of depth. The depth is then decoded via a pretrained VAE~\cite{kingma2013auto}, achieving accurate depth estimates. Thanks to the visual priors of pre-trained diffusion models~\cite{sd}, sharper depth edges are obtained. Later works such as Depth Crafter~\cite{hu2024depthcraftergeneratingconsistentlong} and Depth Any Video~\cite{yang2024depthvideoscalablesynthetic} extend this approach to video depth generation tasks. However, relative depth estimation still faces significant limitations in downstream tasks such as environmental perception and 3D reconstruction~\cite{3dgs,nerf,blendedmvs} due to the absence of absolute scale information.

\subsection{Metric Depth Estimation}

Compared to relative depth estimation, metric depth estimation~\cite{p3depth,ait,dpt,zoedepth,zerodepth,metric3d,metric3dv2} has greater application value due to the presence of precise scale information. Some approaches~\cite{metric3d,metric3dv2,depth_any_camera} consider the uncertainty in depth caused by camera intrinsics and introduce camera intrinsic parameters to transform the perspective image into an intrinsic-independent space for depth estimation, thereby mitigating the uncertainty introduced by camera intrinsics. However, this requirements on camera intrinsics limit the applicability of such methods, such as depth estimation for images generated by AI or images with unknown camera intrinsics. 

Therefore, some methods explore monocular metric depth estimation from a single image without requiring camera intrinsics~\cite{adabins,localbins,zoedepth,depthanythingv1,newcrfs,bts}. While these methods achieve strong in-domain performance by overfitting to specific datasets, they typically require manually setting the maximum truncation depth for each domain. Due to limited training diversity, they struggle to generalize across indoor and outdoor scenes and lack zero-shot capability, resulting in lower accuracy for unseen environments.
Additionally, some works~\cite{scaledepth} decouple the relative depth map prediction from the absolute metric scale prediction, combining them to produce the final metric depth map. Some studies~\cite{localbins,adabins,iebins} argue that direct depth regression across a large depth range is challenging, and thus, manually set maximum depths in a single scene, then, the range from 0 to the maximum depth is adaptively divided into different depth bins. Each pixel’s depth value is classified, and the classification results are weighted to obtain smooth depth results. However, due to the manual setting of the maximum depth, these methods often face degraded generalization across datasets or across indoor and outdoor scenes. 
In this paper, we propose a sliding anchor-based representation to normalize depth from zero to infinity in a unified manner, eliminating the need for manually defining the scene’s maximum depth. This enables our method to scale to larger datasets and achieve improved generalization.

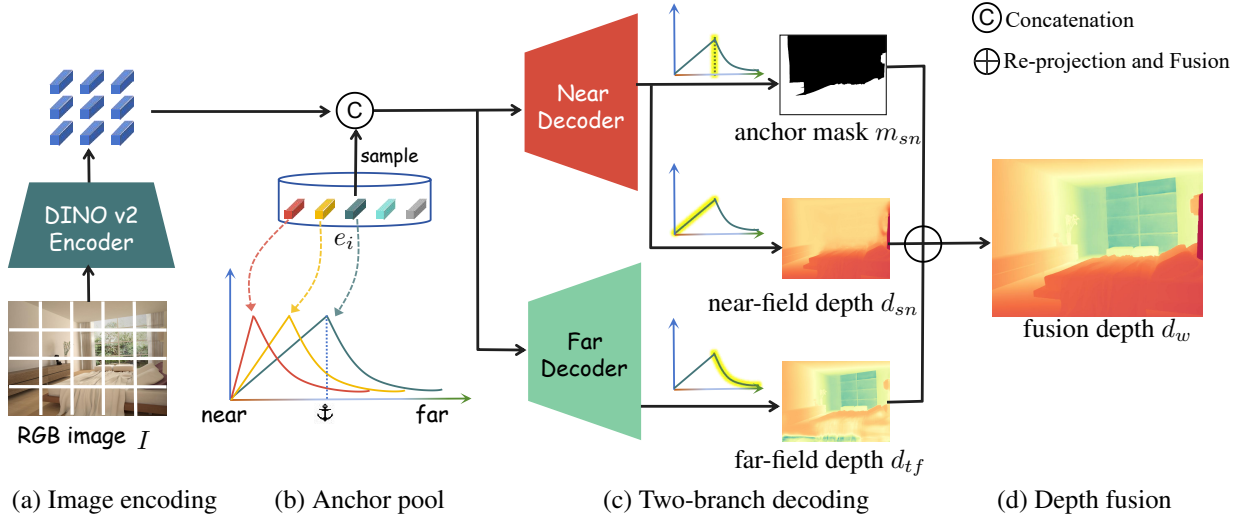
\begin{figure*}[t]
  \centering
   \begin{overpic}[width=\linewidth]{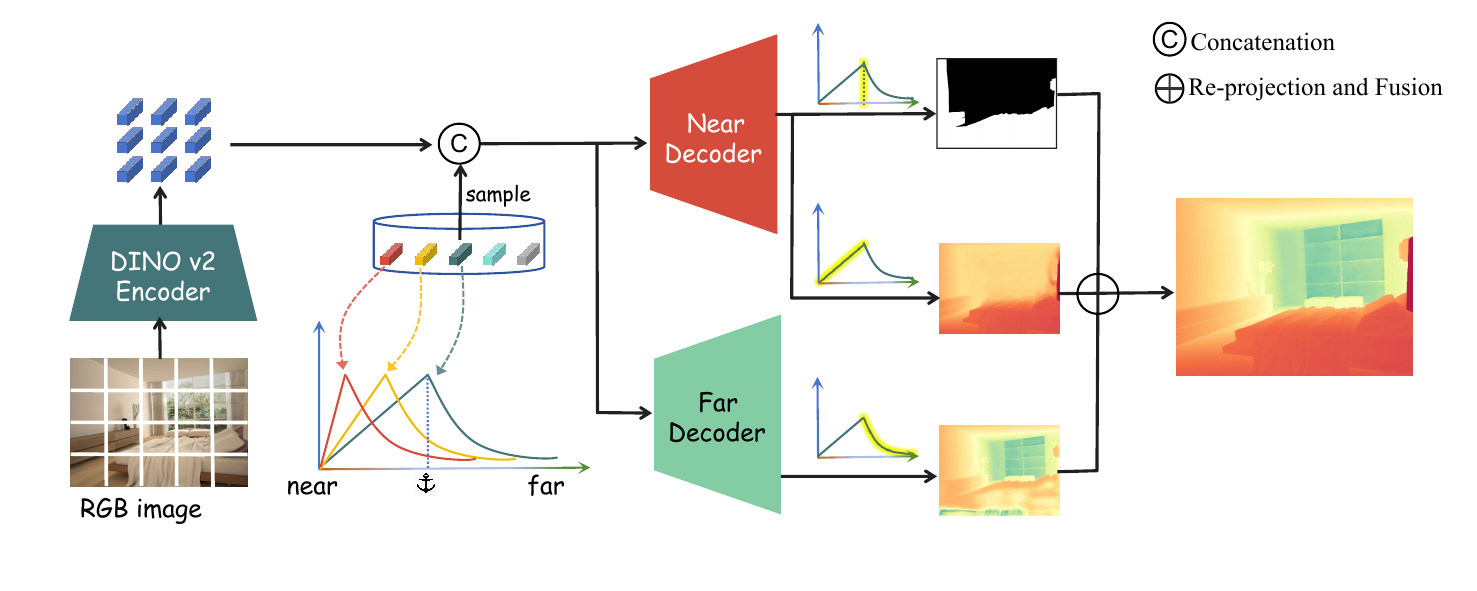}
        \put(14,4.5){ \(I\) }
        \put(29,20){ \( e_i \)}
 
        \put(60,28){anchor mask \( m_{sn} \)}
        \put(58,15){near-field depth \( d_{sn} \)}
        \put(60,3){far-field depth \( d_{tf} \)}
        \put(82,13){fusion depth \( d_{w} \)}

        \put(5,0){(a) Image encoding}
        \put(25,0){(b) Anchor pool}
        \put(50,0){(c) Two-branch decoding}
        \put(80,0){(d) Depth fusion}
   \end{overpic}

    \caption{
    \textbf{Method Overview}. Given an input image, we first employ a large-scale image encoder to extract latent features, as illustrated in (a). Next, these latent features, combined with the sampled anchor depth from the anchor pool, as shown in (b), are fed into a two-branch decoder. Here, the anchor represents a boundary between near and far, and is divided at the pixel level through the anchor mask $m_{sn}$. During training, all different anchors have a chance to be randomly selected from the pool.
    Then the two-branch decoder predicts \neardepth $d_{sn}$, anchor mask $m_{sn}$, and \fardepth $d_{tf}$, as depicted in (c). Finally, the two depth representations are fused using the mask to generate the final complete depth prediction, as demonstrated in (d). 
    }
   \label{fig:method_overview}
   \vspace{-18pt}
\end{figure*}

\section{Method}
\label{sec:methon}

\subsection{Overall Framework}
In this section, we describe our sliding anchor-based approach for monocular metric depth estimation. 
We first introduce the key idea behind the anchor-based depth representation (Sec. \ref{sec:Anchor-based-Representation}), which builds on a pretrained image encoder and a two-branch decoding network and enables seamlessly processing of the depth in the scene from zero to infinity. 
Then, we present the design of sliding anchor {(Sec. \ref{sec:Sliding_Anchor})} to enable more accurate and scalable depth estimation across diverse scenes. 
Finally, we outline the training loss functions used to optimize the model {(Sec. \ref{sec:training_loss})}. 
An overview of our method is shown in Fig. \ref{fig:method_overview}.

\subsection{Anchor-based Depth Representation}\label{sec:Anchor-based-Representation}
Our approach is driven by the observation that metric depth estimation across diverse environments faces a significant challenge: the variation in maximum depth across different scenes. 

To address this challenge, we propose a sliding anchor-based depth representation that dynamically adapts to the scene's depth scale. 

Given an input image \(I\), the first step of our method is to utilize a large image encoder, to extract latent features from the input image. These features serve as a high-level representation of the image, capturing important information about the scene. Then, the encoded latent features, along with sampled anchor embedding, are processed by two decoding branches: (1) \emph{\neardepth} and (2) \emph{\fardepth}, which can formulate a complete depth prediction for the input image.
Furthermore, the anchor depth dynamically adjusts along the depth axis, allowing the model to adapt to different scene scales, ensuring more flexible and accurate depth estimation across diverse environments. Next, we will introduce each components one by one. 

\paragraph{Scaled Near Depth} 

The first branch generates the normalized near-field depth \( d_{\text{sn}}(x, y) \), where \( x, y \) represent the pixel coordinates in the image. This depth lies within the anchor depth \( d_{\text{anchor}} \), and by using \( d_{\text{anchor}} \) as a normalization factor, the depth values in this region can be linearly scaled to the range \( d_{\text{sn}} \in [0,1] \).  
We define this normalization as:  
\vspace{-2pt}
\begin{equation}
d_{sn}(x, y) = \frac{d(x, y)}{d_{anchor}}, 0\leq d(x, y) \leq d_{anchor} 
\end{equation}
\vspace{-2pt}
where \(d(x, y)\) is the true depth at pixel \((x, y)\).

During training, we directly supervise the predicted near-field depth \( d_{sn}(x, y) \) using the ground truth (GT) depth \( \overline{d}_{sn}(x, y) \) in the corresponding near-field regions. However, beyond the reference anchor depth, ground truth supervision is absent, making predictions in these areas unreliable and unstable. This lack of supervision leads to ambiguities in distinguishing valid regions of the scaled near-depth representation during inference, reducing the model’s robustness in generalization. 
Thus, we introduce an additional mask prediction head in this branch, as shown in Fig. \ref{fig:method_overview} (c). Specifically, at the final layer of this branch, before the final depth prediction, we add a linear projection layer that maps the feature into a binary mask. This binary mask indicates the valid areas of the scaled depth prediction: the valid areas are assigned a value of 1 (true), and the invalid areas are assigned a value of 0 (false). By incorporating this mask prediction, the model can effectively differentiate between valid and invalid depth regions, leading to more stable and reliable depth estimation. Formally, we define the predicted mask \( m_{sn}(x,y) \) for the scaled near-depth branch as:
\vspace{-2pt}
\begin{equation}
\overline{m}_{sn}(x, y) = \mathbb{I}(\overline{d}(x, y) \leq d_{anchor})
\end{equation}
\vspace{-2pt}
where \( \mathbb{I}(\cdot) \) is the indicator function, \( \overline{d}(x, y) \) is ground truth depth and \( \overline{m}_{sn}(x, y) \) is GT mask.  

By incorporating this mask prediction mechanism, we ensure that both branches focus on the most reliable parts of the scene, resulting in more accurate and robust depth predictions. 

\paragraph{Tapered Far Depth} 
The second branch generates the tapered far-field depth \( d_{\text{tf}}(x, y) \), which captures depth information beyond the anchor depth. The corresponding depth range is \([d_{\text{anchor}}, \infty]\). To normalize these far-field depths into a consistent range, we apply an exponential normalization function with a negative exponent, ensuring that the anchor depth is mapped to 1, while far depths gradually decay to 0 in a smooth and continuous manner. 

The normalization function is defined as:  
\vspace{-2pt}
\begin{equation}
d_{\text{tf}}(x,y) = e^{-k(d(x,y)-d_{\text{anchor}})},  d(x, y) \geq d_{anchor} 
\end{equation}
\vspace{-2pt}
where \( k \) is a hyperparameter that controls the rate of depth attenuation beyond the reference anchor depth. In our implementation, we set \( k = 0.025 \) to achieve a smooth and stable depth transition.  

\paragraph{Depth Re-projection and Fusion}
After the two branches generate normalized depth representations, they are reprojected into real-world depth values and fused to obtain a complete depth prediction in real-world scale, denoted as $d^w(x,y)$. First, we compute the inverse transformations of $d_{sn}$ and $d_{tf}$ to get $d^w_{sn}$ and $d^w_{tf}$ in real-world scale using the reference anchor depth \(d_{\text{anchor}}\) as the scaling factor. 
\vspace{-2pt}
\begin{equation}
d_{sn}^{w}(x,y) = d_{sn}(x, y) \cdot d_{anchor}
\end{equation}
\vspace{-1pt}
\begin{equation}
d_{tf}^{w}(x,y) =  \frac{-\ln d_{tf}(x, y)}{k} + d_{anchor}
\end{equation}
Then, these two depth components are fused using the mask $m_{sn}(x, y)$, which determines the valid regions for each branch.
\begin{equation}
d^w(x, y) = m_{sn}(x, y) \cdot d_{sn}^w(x, y)  + (1 - m_{sn}(x, y)) \cdot d_{tf}^{w}
\end{equation}

\subsection{Sliding Anchor}\label{sec:Sliding_Anchor}
The core idea of our representation is to use a reference depth as an anchor to normalize depth values from zero to infinity in a unified manner. To achieve robust generalization across varying scene scales, we construct an anchor pool that spans from near to far, consisting of a set of learnable embeddings:  
\vspace{-2pt}
\begin{equation}
E_{\text{anchor}} = \{ e_1, e_2, ..., e_N \}
\end{equation}

where \( e_i \) represents the embedding corresponding to a specific anchor depth \( d_{\text{anchor}, i} \), and \( N \) is the total number of anchor depths in the pool.  

During training, we randomly sample an anchor depth \( d_{\text{anchor}} \) for each input image and use its corresponding embedding \( e_{\text{anchor}} \) to modulate the depth normalization process. This approach allows the model to learn how to adjust depth predictions based on varying scene scales, significantly improving generalization across diverse environments.  
At inference time, the choice of anchor embedding allows us to control the network's focus on different depth ranges. Specifically:  (1) Near-field embedding: If a smaller anchor depth is selected, the network prioritizes closer objects, yielding higher resolution depth estimates for the near-field. This is because a smaller \( d_{\text{anchor}} \) increases the effective resolution within the [0,1] normalized range.  
(2) Far-field embedding: If a larger anchor depth is used, the model extends its focus to distant objects, better capturing depth variations in far regions.  

By leveraging anchor embeddings, our approach enables the model to adaptively adjust depth estimation across scenes with different depth scales, improving accuracy and flexibility in real-world applications.  

\begin{table*}[!htb]
\footnotesize
\centering
\setlength{\tabcolsep}{3pt} 
\renewcommand{\arraystretch}{0.9}
\begin{tabular}
{@{}
    l@{\hspace{6pt}}
    c@{\hspace{6pt}}c@{\hspace{3pt}}c@{\hspace{3pt}}|
    c@{\hspace{6pt}}c@{\hspace{3pt}}c@{\hspace{3pt}}|
    c@{\hspace{6pt}}c@{\hspace{3pt}}c@{\hspace{3pt}}|
    c@{\hspace{6pt}}c@{\hspace{3pt}}c@{\hspace{3pt}}
    @{}}
\toprule
 & \multicolumn{3}{c|}{\textbf{iBims}} & \multicolumn{3}{c|}{\textbf{DIODE Indoor}}   & \multicolumn{3}{c|}{\textbf{DIODE Outdoor}} & \multicolumn{3}{c}{\textbf{SYNTHIA}}\\

Method  & $\delta_{1}$\,$\uparrow$ & REL\,$\downarrow$ & RMSE\,$\downarrow$& $\delta_{1}$\,$\uparrow$ & REL\,$\downarrow$ & RMSE\,$\downarrow$ & $\delta_{1}$\,$\uparrow$ & REL\,$\downarrow$ & RMSE\,$\downarrow$ & $\delta_{1}$\,$\uparrow$ & REL\,$\downarrow$ & RMSE\,$\downarrow$  \\ 
\midrule
AdaBins~\cite{adabins}       & 0.555 & 0.212  & 0.901   & 0.174 & 0.443  & 1.963    & 0.161 & 0.863 & 10.35& 0.832 & 0.350 & 6.271  \\
BTS~\cite{bts}               & 0.538 & 0.231  & 0.919   & 0.210 & 0.418  & 1.905    & 0.171 & 0.837 & 10.48 & 0.863 & 0.785 & 4.920   \\
LocalBins~\cite{localbins}   & 0.558 & 0.211  & 0.880   & 0.229 & 0.412  & 1.853    & 0.170  &  {0.821} & 10.27 & 0.901 & 0.720 & 5.707   \\
NeWCRFs~\cite{newcrfs}       & 0.548 & 0.206  & 0.861   & 0.187 & 0.404  & 1.867    & 0.176 & 0.854 & 9.228 &0.923 & 0.468 & 5.934 \\
ZoeDepth~\cite{zoedepth}     & \underline{0.612} & \underline{0.185} & \underline{0.732}  & 0.247 &  0.371 & 1.842  & \underline{0.269} & 0.852 &\underline{6.898} & 0.912 & 0.413& 4.762  \\
DAV2~\cite{depthanythingv2}  & 0.512 & 0.243  & 0.848   & 0.311 & 1.511  & 6.774   & 0.192 & 1.435 & 10.14 & \underline{0.936} & \underline{0.325} & 4.934  \\
\midrule
ZoeDepth-NK~\cite{zoedepth}  & 0.588 & 0.192 & 0.830 &\underline{0.386} & \underline{0.331} & \underline{1.598}  &  0.208 & \underline{0.757} & 7.569 & 0.902 & 0.824 & \underline{4.274}  \\
\textbf{ours}                & \textbf{0.910} & \textbf{0.111} & \textbf{0.409}    & \textbf{0.446} & \textbf{0.279} & \textbf{0.180}  & \textbf{0.383} & \textbf{0.656} & \textbf{4.836}  & \textbf{0.966} & \textbf{0.153} & \textbf{3.842} \\
\bottomrule
\end{tabular}
\vspace{-6pt}
\caption{\textbf{Zero-shot generalization on indoor and outdoot datasets.} 
Our model achieves the best or near-best performance without the need to switch model parameters, demonstrating the generalization capability across all datasets. 
Note that both our method and ZoeDepth-NK use a single unified model for inference across all datasets, whereas other methods employ separate models for indoor and outdoor scenes.
}
\label{tab:zero-shot-all}
\vspace{-15pt}
\end{table*}

\subsection{Training}\label{sec:training_loss}

During training, we optimize the model using three distinct loss components derived from both branches, including a depth loss for each branch and a mask loss to distinguish valid regions.

\paragraph{Scaled Near Depth Loss}
The scaled near depth loss, denoted as $\mathcal{L}_{sn}$, 
measures the difference between the predicted scaled near depth $d_{sn}(x, y)$ and the ground truth $\overline{d}_{sn}(x, y)$. We use the $L_2$ loss to compute this difference:
\begin{equation}
\mathcal{L}_{sn} = \sum_{x, y} \left( d_{sn}(x, y) - \overline{d}_{sn}(x, y) \right)^2 \cdot \overline{m}_{sn}(x,y)
\end{equation}

\paragraph{Tapered Far Depth Loss}
Similarly, the tapered far depth loss, denoted as \( \mathcal{L}_{tf} \), measures the difference between the predicted depth \( d_{tf}(x, y) \) and the ground truth depth \( \overline{d}_{tf}(x, y) \):
\vspace{-2pt}
\begin{equation}
\mathcal{L}_{tf} = \sum_{x, y} \left( d_{tf}(x, y) - \overline{d}_{tf}(x, y) \right)^2 \cdot (1 - \overline{m}_{sn}(x,y))
\end{equation}
\paragraph{Scaled Near Depth Mask Loss}
The mask loss for the scaled depth branch, denoted as \( \mathcal{L}_{mask} \), is computed using binary cross-entropy (BCE) loss between the predicted mask \( m_{sn}(x, y) \) and the ground truth \( \overline{m}_{sn}(x, y) \). The BCE loss can be expressed as:
\begin{equation}
\mathcal{L}_{mask} = \sum_{x, y} \text{BCE}(m_{sn}(x, y), \overline{m}_{sn}(x, y))
\end{equation}

\paragraph{Total Loss}
The total loss for the model is the sum of the three losses from both branches:
\vspace{-2pt}
\begin{equation}
\mathcal{L}_{\text{total}} = \lambda_{sn}*\mathcal{L}_{sn} + \lambda_{tf} * \mathcal{L}_{tf} + \lambda_m * \mathcal{L}_{mask}
\end{equation}
where $\lambda_{sn}$, $\lambda_{tf}$, and $\lambda_m$ are set to 1.0, 1.0, and 0.05, respectively, to balance the contributions of the three corresponding loss components, respectively.

This comprehensive loss function allows the model to jointly optimize for accurate depth predictions and reliable mask predictions, ensuring that the model focuses on valid regions of the depth map while making robust predictions across varying scene scales.

\section{Experiments}
\label{sec:exp}

\subsection{Implementation Details}
Our model architecture consists of an image encoder and two decoding branches. The first branch predicts normalized real depth, while the second branch predicts normalized reversed depth. Both branches are adapted from the Dense Prediction Transformer (DPT) \cite{dpt} head to align with our sliding anchor-based approach.
For the image encoder, we use DINOv2~\cite{dinov2}, initialized with pretrained weights from DepthAnythingV2 large \cite{depthanythingv2}. The DPT heads are randomly initialized to learn depth predictions tailored to our method. 
The training is conducted using the AdamW~\cite{loshchilov2017decoupled} optimizer with a learning rate of \(5\times10^{-6}\), running on 8 NVIDIA H100 GPUs for a total of 205K steps. 
And the model is pretrained on a combination of diverse datasets covering a wide range of scene scales. Please refer to the supplemental material for detailed information on the datasets used.
\begin{table}[t]
\scriptsize
\captionsetup{font=footnotesize}
\centering
\begin{subtable}{0.5\textwidth}
    \centering
    \setlength\tabcolsep{1.4mm}
    \renewcommand{\arraystretch}{1.05}
    \begin{tabular}{rcccccc}
    \toprule
    \multirow{2}{*}{Method} & \multicolumn{3}{c}{\emph{Higher is better} $\uparrow$} & \multicolumn{3}{c}{\emph{Lower is better} $\downarrow$} \\

    \cmidrule(lr){2-4}\cmidrule(lr){5-7}
    
    ~ & $\delta_1$ & $\delta_2$ & $\delta_3$ & AbsRel & RMSE & log10 \\
    
    \midrule 
    
    AdaBins-N~\cite{adabins} & 0.903 & 0.984 & 0.997 & 0.103 & 0.364 & 0.044 \\
    NeWCRFs-N~\cite{newcrfs}   & 0.954 & 0.981 & 0.997 & 0.113 & 0.394 & 0.083 \\
    DPT~\cite{dpt}  & 0.904 & 0.988 & 0.998 & 0.110 & 0.357 & 0.045 \\

    P3Depth~\cite{p3depth}  & 0.898 & 0.981 & 0.996 & 0.104 & 0.356 & 0.043 \\
    
    SwinV2~\cite{swinv2}  & 0.949 & 0.994 & 0.999 & 0.083 & 0.287 & 0.035 \\
    
    AiT~\cite{ait} & 0.954 & 0.994 & 0.999 & 0.076 & 0.275 & 0.033 \\

    VPD~\cite{vpd}& 0.964 & 0.995 & 0.999 & 0.069 & 0.254 & 0.030 \\

    IEBins~\cite{iebins}  & 0.936 & 0.992 & 0.998 & 0.087 & 0.314 & 0.038 \\
    
    ZoeDepth-N~\cite{zoedepth} & 0.955 & 0.995 & 0.999 & 0.075 & 0.269 & 0.032 \\
    DAV1-N~\cite{depthanythingv1} &  \underline{0.984} & \underline{0.998} & \textbf{1.000} & \underline{0.056} & \underline{0.205} & \underline{0.024} \\
    \midrule
    ZoeDepth-NK~\cite{zoedepth} & 0.952 & 0.995 & 0.999 & 0.077 & 0.280 & 0.033 \\

    Ours  & \textbf{0.986} & \textbf{0.998} & \underline{0.999} & \textbf{0.049} & \textbf{0.183} & \textbf{0.021} \\
    \bottomrule
    \end{tabular}
    \caption{We compare our method against state-of-the-art (SOTA) monocular depth estimation models on NYU-D dataset}
    \label{tab:metric_nyu}
\end{subtable}
\hfill
\begin{subtable}{0.52\textwidth}
    \centering
    \setlength\tabcolsep{1.3mm}
    \renewcommand{\arraystretch}{1.048}
    \begin{tabular}{rcccccc}
    \toprule
    \multirow{2}{*}{Method} & \multicolumn{3}{c}{\emph{Higher is better} $\uparrow$} & \multicolumn{3}{c}{\emph{Lower is better} $\downarrow$} \\

    \cmidrule(lr){2-4}\cmidrule(lr){5-7}
    
    ~ & $\delta_1$ & $\delta_2$ & $\delta_3$ & AbsRel & RMSE & log10 \\

    \midrule
    
    AdaBins-K~\cite{adabins} & 0.944 & 0.991 & 0.998 & 0.071 & 3.039 & 0.031 \\
    
    P3Depth~\cite{p3depth} & 0.953 & 0.993 & 0.998 & 0.071 & 2.842 & 0.103 \\
    
    NeWCRFs-K~\cite{newcrfs}& 0.974 & 0.997 & 0.999 & 0.052 & 2.129 & 0.079 \\
    
    SwinV2~\cite{swinv2} & 0.977 & 0.998 & 1.000 & 0.050 &\underline{1.966} & 0.075\\

    NDDepth~\cite{nddepth} &\underline{0.978} & 0.998 & 0.999 & 0.050 & 2.025 & 0.075 \\

    IEBins~\cite{iebins}& 0.976 & 0.997 & 0.999 & \underline{0.048} & 2.044 & 0.076 \\

    DAV1-K~\cite{depthanythingv1}  &\textbf{0.982} &\underline{0.998} & \underline{1.000} & \textbf{0.046} & \textbf{1.896} & \textbf{0.024} \\
    \midrule
    ZoeDepth-NK~\cite{zoedepth} & 0.971 & 0.996 & 0.999 & 0.054 & 2.281 & 0.082 \\

    Ours & 0.976 & \textbf{0.998} & \textbf{1.000} & 0.052 & 2.281 & \underline{0.031}  \\

    \bottomrule
    \end{tabular}
    \caption{We compare our method against state-of-the-art (SOTA) monocular depth estimation models on KITTI dataset}
    \label{tab:metric_kitti}
\end{subtable}
    \caption{
    \textbf{Quantitative in-domain metric depth estimation}. 
    (a) NYU-D for indoor and (b) KITTI for outdoor. All compared methods use the encoder size close to ViT-L.
    Each sub-table consists of two blocks: the first block shows results from methods fine-tuned on a specific domain (indoor or outdoor), while the second block presents results from methods fine-tuned jointly on both domains.
    The model names suffixed with ``N'' indicate models specifically optimized for NYU, while those with ``K'' denote models optimized for KITTI.
    }
    \label{tab:metric_nyu_kitti}
\end{table}

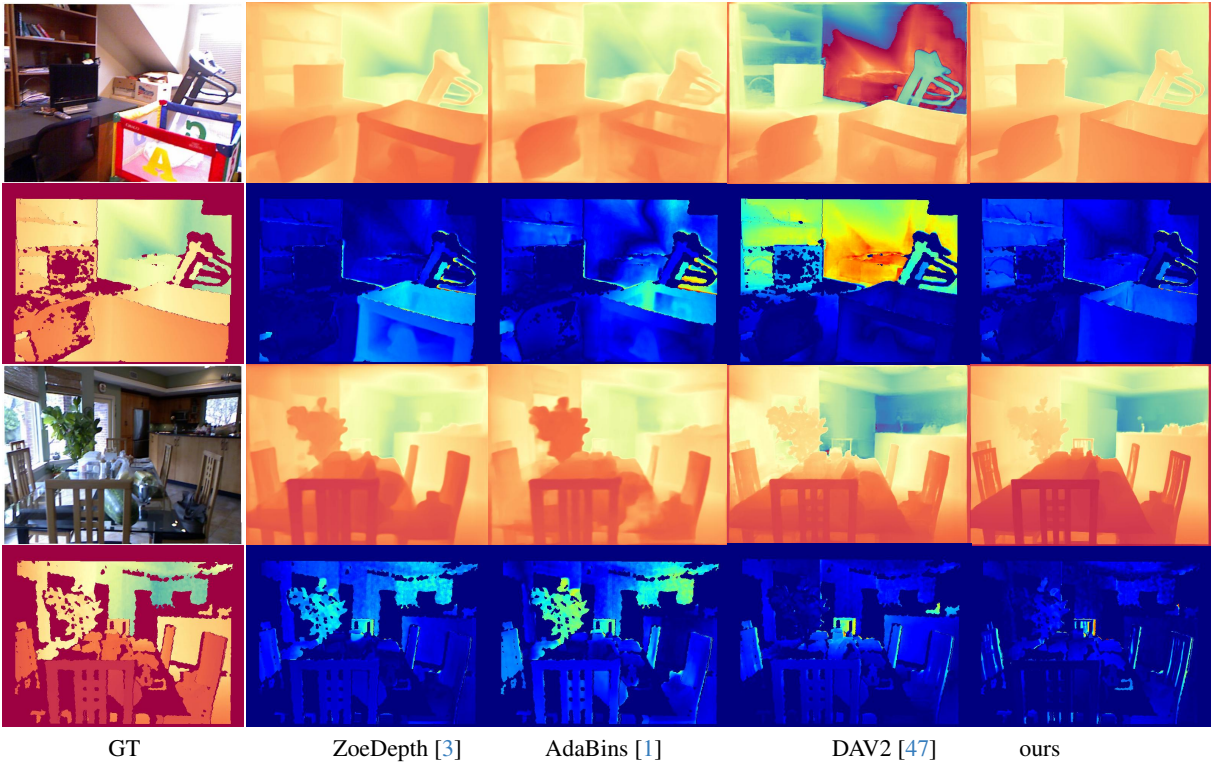
\begin{figure*}[t]
  \centering
  \begin{overpic}[width=0.95\linewidth]{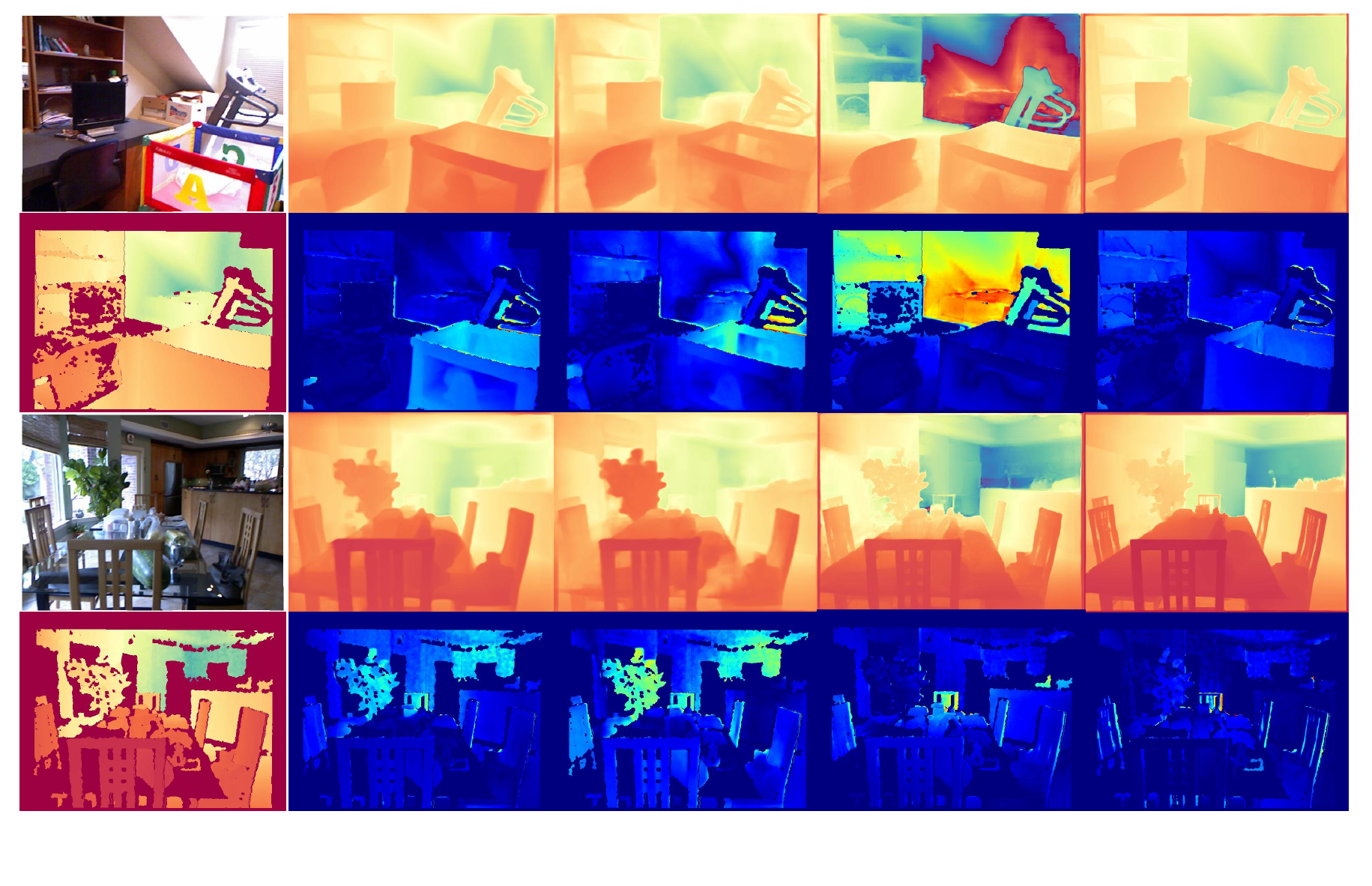}
  \put(10,3){\small GT}
   \put(28,3){\small ZoeDepth~\cite{zoedepth}}
    \put(45,3){\small 
 AdaBins~\cite{adabins}}
     \put(68,3){\small 
 DAV2~\cite{depthanythingv2}}
      \put(83,3){\small  ours}
  \end{overpic}
  \vspace{-18pt}
   \caption{\textbf{Qualitative comparisons of depth predictions on the indoor dataset NYU}. We show both depth maps and corresponding error maps. When dealing with large-scale and long-distance indoor scenes, our framework achieves better absolute depth recovery.
   }
   \label{fig:comp_depth_pred_in}
   \vspace{-10pt}
\end{figure*}

\subsection{Comparisons}
\paragraph{Zero-shot Generalization}
To test the generalization performance of different models in zero-shot indoor and outdoor scenarios, we conducted performance comparisons in 4 datasets: 2 indoor datasets (iBims~\cite{ibims} and DIODE~\cite{diode} Indoor) and 2 outdoor datasets
(DIODO~\cite{diode} Outdoor and SYNTHIA~\cite{synthia}). 
As shown in Table \ref{tab:zero-shot-all}, our model achieves the best generalization performance across all datasets. This further demonstrates the effectiveness of our design.

\paragraph{Comparison in NYU and KITTI}
Since most metric depth estimation methods are fine-tuned separately on NYU-V2~\cite{nyud} and KITTI~\cite{kitti}, these datasets serve as the primary benchmarks for evaluating indoor and outdoor performance. 
We finetune the pretrained model on these two datasets for 70 epochs and compare our method against state-of-the-art (SOTA) monocular depth estimation models on NYU-V2 (indoor) and KITTI (outdoor). 
Because other methods impose a maximum depth threshold, we limit our evaluation depth range to 10m for NYU-V2 and 80m for KITTI to ensure a fair comparison. However, it is important to note that our model's effective estimation range is not constrained by these limits.

As shown in Table \ref{tab:metric_nyu_kitti}, our single model outperforms the baselines on both indoor and outdoor datasets (i.e., the second block in each sub-table), even when some baselines are fine-tuned on each dataset individually (i.e., the first block in each sub-table). This improvement is attributed to our joint training across multiple datasets, which enhances our model’s generalization ability across diverse depth distributions.

The qualitative visualization experiments can be seen in Fig. ~\ref{fig:comp_depth_pred_in} and ~\ref{fig:comp_depth_pred_outdoor} 
in indoor scenes and outdoor street scenes, respectively. It can be observed that our approach not only captures details more accurately and achieves the lowest overall error but also demonstrates the farthest prediction range on the KITTI dataset. Additionally, it effectively handles complex scenarios such as the sky.

\subsection{Analysis}
\paragraph{Ablation}
To ablate the significance of our design choices, we conduct experiments using training data from vkitti~\cite{vkitti} and HyperSim~\cite{hypersim}, respectively. We test three different settings to assess the impact of each component: 
(2) Without mask head: In this setup, we use the reference anchor depth to truncate the predictions and fuse the predictions naively, without the additional mask prediction head. 
(3) Our full setting: This is the full configuration, including both branches with mask prediction and anchor depth injection. 
Table \ref{tab:ablation_study} presents the quantitative results for all three settings. 
Our full setting achieves the best performance across all metrics.

\begin{figure*}[t]
  \centering
  \begin{overpic}[width=\linewidth]
  {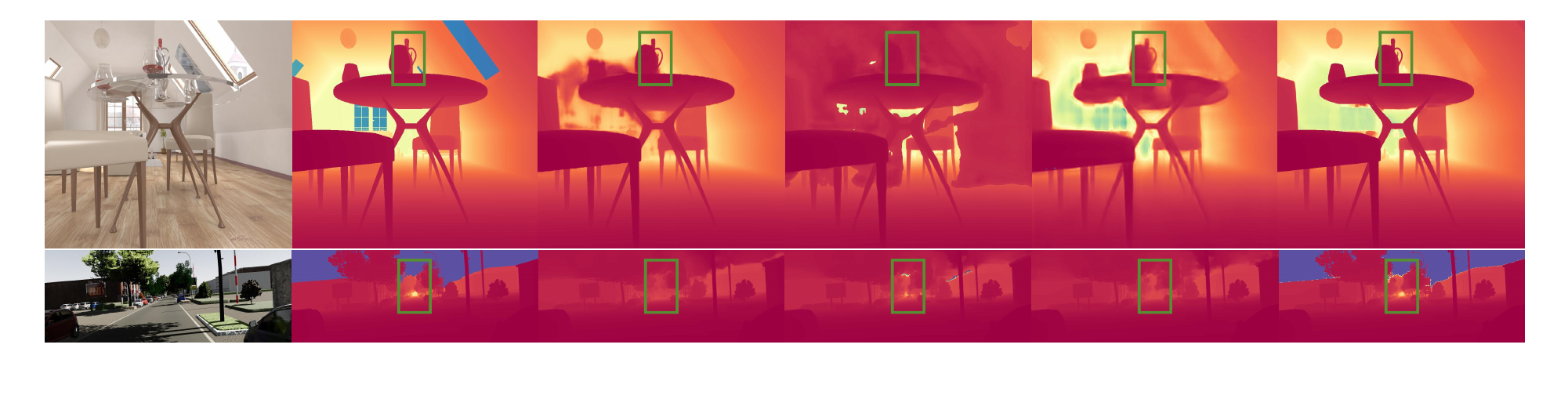}
    \put(8,3){\small (a) image}
    \put(24,3){\small (b) GT}
    \put(38,3){\small (c) one head}
    \put(55,3){\small (d) w/o mask}
    \put(70,3){\small (e) fix anchor}
    \put(87,3){\small (f) ours}
  \end{overpic}
  \vspace{-32pt}
   \caption{\textbf{Qualitative comparisons of different ablation settings}. Compared with the baseline settings (c), our full setting (f) allows for effective observing further distances (e.g., sky in the second row). And the anchor mask-based fusion strategy ensures seamless stitching of near and far depths (d) and higher depth fidelity in near-range indoor scenes (e) in indoor scenes.
   }
   \vspace{-15pt}
   \label{fig:ablation}
\end{figure*}

Fig.~\ref{fig:ablation} illustrates the qualitative performance. 
To highlight the advantage of our two-branch design, we include an additional setting named One-head, which uses a single branch trained on all data with a fixed maximum depth of 80 meters to accommodate both indoor and outdoor environments. 
As shown in Fig.~\ref{fig:ablation} (c) (second row), the model fails to predict depths beyond the predefined maximum depth.
In Fig.\ref{fig:ablation}(d) (first row), removing the mask head cannot produce accurate fusion of the two depth branches. As illustrated in Fig.\ref{fig:ablation}(e), using a fixed anchor results in the loss of fine-grained details.
These results demonstrate the effectiveness of the sliding anchor-based depth representation and the importance of the mask prediction mechanism.

\begin{table}
    \resizebox{0.98\linewidth}{!}{ 
    \begin{tabular}{lccccc}
    \toprule
    Model setting   & $\delta_{1}$\,$\uparrow$ & $\delta_{2}$\,$\uparrow$  & REL\,$\downarrow$ & RMSE\,$\downarrow$ & log10\,$\downarrow$ \\ 
    \midrule
    fix-anchor & 0.701 & 0.917 & 0.200 & 2.744 & 0.085 \\
    w/o mask head & 0.341 & 0.813 & 0.563 & 10.324 & 0.321\\
    Ours (full) & \textbf{0.734} & \textbf{0.935} & \textbf{0.189} & \textbf{2.616} & \textbf{0.071}  \\
    \bottomrule
    \end{tabular}
    \vspace{-6pt}
    }
    \caption{\textbf{Ablation studies of model design}. 
    We compare the performance using fixed anchors and a fusion strategy without masking separately. Our full setting achieves the best performance.}
    \label{tab:ablation_study}
    \vspace{-10pt}
\end{table}

\begin{figure}[t]
  \centering
  \begin{overpic}[width=1.0\linewidth]{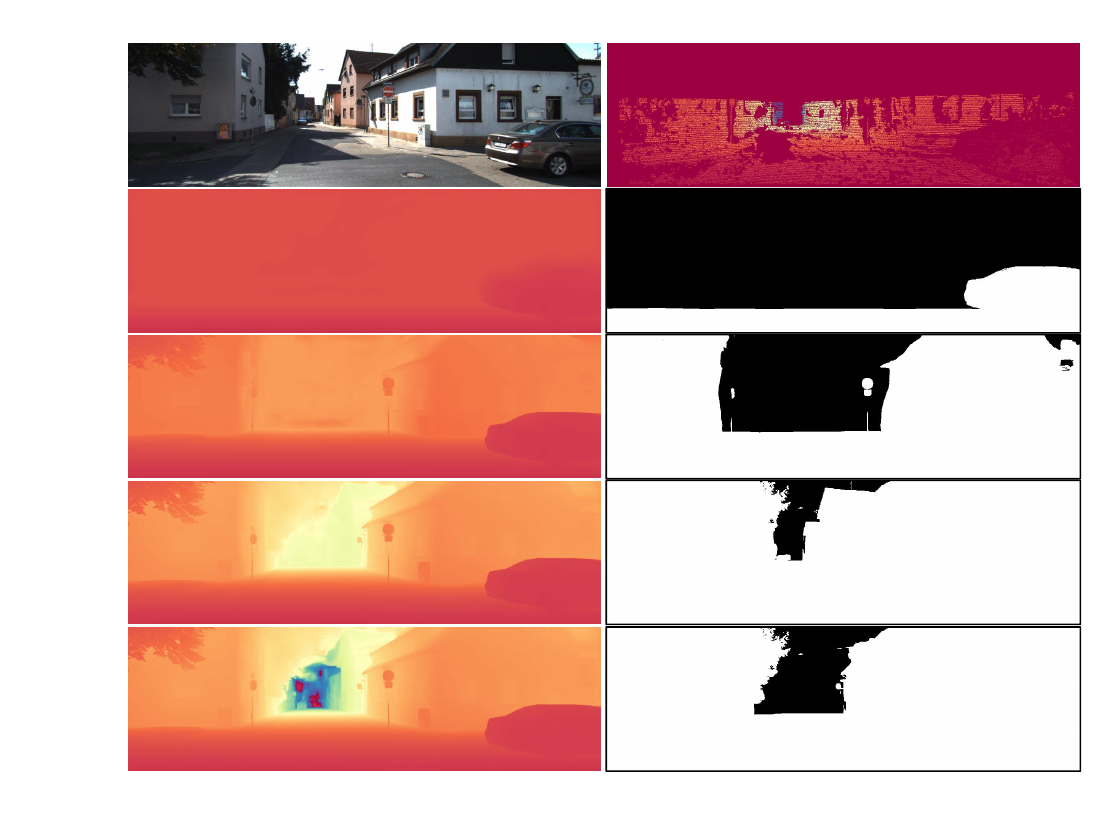}
    \put(20,72){Reference}
    \put(70,72){GT (Lidar)}
    \put(20,0){Near depth output}
    \put(68,0){Fusion mask}
    \put(-3,54){Anchor}
    \put(0,49){10m}
    \put(0,38){20m}
    \put(0,24){40m}
    \put(0,10){80m}
  \end{overpic}
   \caption{\textbf{Qualitative comparisons of different reference anchor depth}.It can be observed that anchors at different distances allow the near head to precisely focus on depths within different ranges and provide accurate anchor depth masks.
   }
   \vspace{-8pt}
   \label{fig:vis_diff_anchor}
\end{figure}
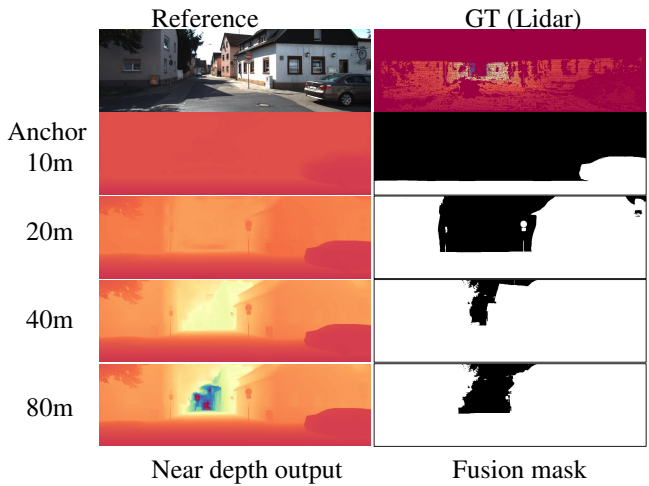
\begin{table}
    \centering
    \resizebox{0.98\linewidth}{!}{
    \begin{tabular}{ccccccc}
    \toprule
    \noalign{\vspace{4pt}}
    ~ & \multicolumn{3}{c}{\textbf{vkitti(outdoor)}} &   & \\
    \hline
    \noalign{\vspace{4pt}}
    & \multicolumn{2}{c}{\textbf{20m}} & \multicolumn{2}{c}{\textbf{80m}} & \multicolumn{2}{c}{\textbf{120m}} \\
    Anchor  & $\delta_{1}$\,$\uparrow$ & REL\,$\downarrow$ & $\delta_{1}$\,$\uparrow$ & REL\,$\downarrow$ & $\delta_{1}$\,$\uparrow$ & REL\,$\downarrow$\\ 
    \noalign{\smallskip}
    \hline\noalign{\smallskip}
    Anchor-1(20m) & \textbf{0.963} & \textbf{0.066} & 0.871 & 0.128 & 0.771 & 0.154 \\
    Anchor-2(40m) &  0.960 & 0.068 & 0.917 & \textbf{0.092} & 0.804 & 0.147\\
    Anchor-3(80m) & 0.953 & 0.071 & \textbf{0.929} & 0.096 & 0.851 & 0.155\\
    Anchor-4(120m) & 0.936 & 0.086 & 0.916 & 0.108 & \textbf{0.908} & \textbf{0.121} \\
    \midrule
    \noalign{\vspace{4pt}}
    ~ & \multicolumn{3}{c}{\textbf{hypersim (indoor)}} &   &  \\
    \hline
    \noalign{\vspace{4pt}}
    & \multicolumn{2}{c}{\textbf{2m}} & \multicolumn{2}{c}{\textbf{4m}} & \multicolumn{2}{c}{\textbf{10m}} \\
    Anchor  & $\delta_{1}$\,$\uparrow$ & REL\,$\downarrow$ & $\delta_{1}$\,$\uparrow$ & REL\,$\downarrow$ & $\delta_{1}$\,$\uparrow$ & REL\,$\downarrow$\\ 
    \hline
    Anchor-1(2m) & \textbf{0.771} & \textbf{0.176}  & 0.760 & \textbf{0.173} & 0.708 & 0.204 \\
    Anchor-2(4m) & 0.762 & 0.182 & \textbf{0.763} & 0.176 & 0.712 & 0.201 \\
    Anchor-3(6m) & 0.759  & 0.193  &0.734 & 0.182&0.716 & 0.198 \\
    Anchor-4(10m) & 0.761 & 0.189 & 0.731 & 0.201 & \textbf{0.735} & \textbf{0.184} \\
    \bottomrule
    \end{tabular}
    }
    \caption{\textbf{Evaluation of different anchor}.
    The first and second rows indicate the dataset and truncation depth for evaluation, respectively, where only the depth regions within the truncation value are evaluated.
    The model achieves optimal performance when the truncation depth aligns with the anchor depth..
    }
    \label{tab:ablation_study_anchor}
    \vspace{-18pt}
\end{table}

\paragraph{Anchor Embeddings}

To evaluate the impact of anchor embeddings, we conduct experiments on the vkitti~\cite{vkitti} dataset, using consistent input data while varying anchor embeddings as depth prediction conditions. Performance is evaluated across different maximum truncation depths.
As shown in Table \ref{tab:ablation_study_anchor}, the model achieves the best results when the truncation depth matches the anchor depth, indicating that it effectively adapts to different anchor embeddings and produces optimal predictions when evaluated within the corresponding depth range.
Notably, the model achieves the highest accuracy with the smallest anchor embedding, demonstrating that a smaller anchor improves depth precision for closer objects. Fig. \ref{fig:vis_diff_anchor} presents a visualization of the impact of different anchor embeddings, illustrating that as the anchor depth increases, the valid prediction region (i.e., the fusion mask) of the scaled depth branch expands accordingly.

\section{Conclusion and Limitations}
\label{sec:conclusion}

In this paper, we proposed a sliding anchor-based metric depth estimation method to address the challenges of scale variation across diverse environments. Our dynamic sliding anchor allows the model to adapt to varying depth scales, ensuring precise predictions in both near and far fields.
Our framework leverages a pretrained DINOv2 encoder and two modified DPT heads for depth prediction while incorporating learnable anchor embeddings to seamlessly encode depth reference information. Additionally, we introduced a mask prediction mechanism to enhance the robust fusion of depth predictions from two branches, improving model stability and generalization across datasets.
Extensive experiments demonstrate that our method outperforms existing approaches on both indoor and outdoor benchmarks, achieving strong generalization without relying on scene-specific assumptions such as fixed maximum depths or prior knowledge of the scene type. Our findings suggest that the proposed sliding anchor-based representation offers a scalable and effective solution for metric depth estimation across a wide range of real-world applications.

\textbf{Limitations} However, Metric-Solver is only implemented for single images, but we plan to integrate monocular video metric depth estimation in the future. Moreover, necessary optimizations are still required in terms of model efficiency and size.

{
    \small
    \bibliographystyle{setup/ieeenat_fullname}
    \bibliography{main}

\begin{thebibliography}{53}
\providecommand{\natexlab}[1]{#1}
\providecommand{\url}[1]{\texttt{#1}}
\expandafter\ifx\csname urlstyle\endcsname\relax
  \providecommand{\doi}[1]{doi: #1}\else
  \providecommand{\doi}{doi: \begingroup \urlstyle{rm}\Url}\fi

\bibitem[Bhat et~al.(2021)Bhat, Alhashim, and Wonka]{adabins}
Shariq~Farooq Bhat, Ibraheem Alhashim, and Peter Wonka.
\newblock Adabins: Depth estimation using adaptive bins.
\newblock In \emph{CVPR}, 2021.

\bibitem[Bhat et~al.(2022)Bhat, Alhashim, and Wonka]{localbins}
Shariq~Farooq Bhat, Ibraheem Alhashim, and Peter Wonka.
\newblock Localbins: Improving depth estimation by learning local distributions.
\newblock In \emph{European Conference on Computer Vision}, pages 480--496. Springer, 2022.

\bibitem[Bhat et~al.(2023)Bhat, Birkl, Wofk, Wonka, and M{\"u}ller]{zoedepth}
Shariq~Farooq Bhat, Reiner Birkl, Diana Wofk, Peter Wonka, and Matthias M{\"u}ller.
\newblock Zoedepth: Zero-shot transfer by combining relative and metric depth.
\newblock \emph{arXiv:2302.12288}, 2023.

\bibitem[Birkl et~al.(2023)Birkl, Wofk, and M{\"u}ller]{midasv31}
Reiner Birkl, Diana Wofk, and Matthias M{\"u}ller.
\newblock Midas v3. 1--a model zoo for robust monocular relative depth estimation.
\newblock \emph{arXiv:2307.14460}, 2023.

\bibitem[Burnett et~al.(2019)Burnett, Samavi, Waslander, Barfoot, and Schoellig]{autotrack}
Keenan Burnett, Sepehr Samavi, Steven Waslander, Timothy Barfoot, and Angela Schoellig.
\newblock autotrack: A lightweight object detection and tracking system for the sae autodrive challenge.
\newblock In \emph{2019 16th Conference on Computer and Robot Vision (CRV)}, pages 209--216. IEEE, 2019.

\bibitem[Cabon et~al.(2020)Cabon, Murray, and Humenberger]{vkitti}
Yohann Cabon, Naila Murray, and Martin Humenberger.
\newblock Virtual kitti 2.
\newblock \emph{arXiv:2001.10773}, 2020.

\bibitem[Eigen et~al.(2014)Eigen, Puhrsch, and Fergus]{eigen2014depth}
David Eigen, Christian Puhrsch, and Rob Fergus.
\newblock Depth map prediction from a single image using a multi-scale deep network.
\newblock In \emph{NeurIPS}, 2014.

\bibitem[Fu et~al.(2018)Fu, Gong, Wang, Batmanghelich, and Tao]{fu2018deep}
Huan Fu, Mingming Gong, Chaohui Wang, Kayhan Batmanghelich, and Dacheng Tao.
\newblock Deep ordinal regression network for monocular depth estimation.
\newblock In \emph{CVPR}, 2018.

\bibitem[Fu et~al.(2024)Fu, Yin, Hu, Wang, Ma, Tan, Shen, Lin, and Long]{geowizard}
Xiao Fu, Wei Yin, Mu Hu, Kaixuan Wang, Yuexin Ma, Ping Tan, Shaojie Shen, Dahua Lin, and Xiaoxiao Long.
\newblock Geowizard: Unleashing the diffusion priors for 3d geometry estimation from a single image.
\newblock \emph{arXiv:2403.12013}, 2024.

\bibitem[Geiger et~al.(2013)Geiger, Lenz, Stiller, and Urtasun]{kitti}
Andreas Geiger, Philip Lenz, Christoph Stiller, and Raquel Urtasun.
\newblock Vision meets robotics: The kitti dataset.
\newblock \emph{The International Journal of Robotics Research}, 2013.

\bibitem[Gui et~al.(2024)Gui, Fischer, Prestel, Ma, Kotovenko, Grebenkova, Baumann, Hu, and Ommer]{depthfm}
Ming Gui, Johannes~S Fischer, Ulrich Prestel, Pingchuan Ma, Dmytro Kotovenko, Olga Grebenkova, Stefan~Andreas Baumann, Vincent~Tao Hu, and Bj{\"o}rn Ommer.
\newblock Depthfm: Fast monocular depth estimation with flow matching.
\newblock \emph{arXiv:2403.13788}, 2024.

\bibitem[Guizilini et~al.(2023)Guizilini, Vasiljevic, Chen, Ambruș, and Gaidon]{zerodepth}
Vitor Guizilini, Igor Vasiljevic, Dian Chen, Rareș Ambruș, and Adrien Gaidon.
\newblock Towards zero-shot scale-aware monocular depth estimation.
\newblock In \emph{ICCV}, 2023.

\bibitem[Guo et~al.(2025)Guo, Garg, Miangoleh, Huang, and Ren]{depth_any_camera}
Yuliang Guo, Sparsh Garg, S~Mahdi~H Miangoleh, Xinyu Huang, and Liu Ren.
\newblock Depth any camera: Zero-shot metric depth estimation from any camera.
\newblock \emph{arXiv preprint arXiv:2501.02464}, 2025.

\bibitem[Hu et~al.(2024{\natexlab{a}})Hu, Yin, Zhang, Cai, Long, Chen, Wang, Yu, Shen, and Shen]{metric3dv2}
Mu Hu, Wei Yin, Chi Zhang, Zhipeng Cai, Xiaoxiao Long, Hao Chen, Kaixuan Wang, Gang Yu, Chunhua Shen, and Shaojie Shen.
\newblock Metric3d v2: A versatile monocular geometric foundation model for zero-shot metric depth and surface normal estimation.
\newblock \emph{arXiv:2404.15506}, 2024{\natexlab{a}}.

\bibitem[Hu et~al.(2024{\natexlab{b}})Hu, Gao, Li, Zhao, Cun, Zhang, Quan, and Shan]{hu2024depthcraftergeneratingconsistentlong}
Wenbo Hu, Xiangjun Gao, Xiaoyu Li, Sijie Zhao, Xiaodong Cun, Yong Zhang, Long Quan, and Ying Shan.
\newblock Depthcrafter: Generating consistent long depth sequences for open-world videos, 2024{\natexlab{b}}.

\bibitem[Kalia et~al.(2019)Kalia, Navab, and Salcudean]{real}
Megha Kalia, Nassir Navab, and Tim Salcudean.
\newblock A real-time interactive augmented reality depth estimation technique for surgical robotics.
\newblock In \emph{2019 international conference on robotics and automation (icra)}, pages 8291--8297. IEEE, 2019.

\bibitem[Ke et~al.(2024)Ke, Obukhov, Huang, Metzger, Daudt, and Schindler]{marigold}
Bingxin Ke, Anton Obukhov, Shengyu Huang, Nando Metzger, Rodrigo~Caye Daudt, and Konrad Schindler.
\newblock Repurposing diffusion-based image generators for monocular depth estimation.
\newblock In \emph{CVPR}, 2024.

\bibitem[Kerbl et~al.(2023)Kerbl, Kopanas, Leimk{\"u}hler, and Drettakis]{3dgs}
Bernhard Kerbl, Georgios Kopanas, Thomas Leimk{\"u}hler, and George Drettakis.
\newblock 3d gaussian splatting for real-time radiance field rendering.
\newblock \emph{TOG}, 2023.

\bibitem[Kingma et~al.(2013)Kingma, Welling, et~al.]{kingma2013auto}
Diederik~P Kingma, Max Welling, et~al.
\newblock Auto-encoding variational bayes, 2013.

\bibitem[Koch et~al.(2018)Koch, Liebel, Fraundorfer, and Korner]{ibims}
Tobias Koch, Lukas Liebel, Friedrich Fraundorfer, and Marco Korner.
\newblock Evaluation of cnn-based single-image depth estimation methods.
\newblock In \emph{ECCVW}, 2018.

\bibitem[Lee and Kim(2019)]{monocular}
Jae-Han Lee and Chang-Su Kim.
\newblock Monocular depth estimation using relative depth maps.
\newblock In \emph{Proceedings of the IEEE/CVF Conference on Computer Vision and Pattern Recognition}, pages 9729--9738, 2019.

\bibitem[Lee et~al.(2019)Lee, Han, Ko, and Suh]{bts}
Jin~Han Lee, Myung-Kyu Han, Dong~Wook Ko, and Il~Hong Suh.
\newblock From big to small: Multi-scale local planar guidance for monocular depth estimation.
\newblock \emph{arXiv preprint arXiv:1907.10326}, 2019.

\bibitem[Liu et~al.(2022)Liu, Hu, Lin, Yao, Xie, Wei, Ning, Cao, Zhang, Dong, et~al.]{swinv2}
Ze Liu, Han Hu, Yutong Lin, Zhuliang Yao, Zhenda Xie, Yixuan Wei, Jia Ning, Yue Cao, Zheng Zhang, Li Dong, et~al.
\newblock Swin transformer v2: Scaling up capacity and resolution.
\newblock In \emph{CVPR}, 2022.

\bibitem[Loshchilov and Hutter(2017)]{loshchilov2017decoupled}
Ilya Loshchilov and Frank Hutter.
\newblock Decoupled weight decay regularization.
\newblock \emph{arXiv preprint arXiv:1711.05101}, 2017.

\bibitem[Mertan et~al.(2022)Mertan, Duff, and Unal]{SIDEreviewMERTAN2022103441}
Alican Mertan, Damien~Jade Duff, and Gozde Unal.
\newblock Single image depth estimation: An overview.
\newblock \emph{Digital Signal Processing}, 123:\penalty0 103441, 2022.

\bibitem[Michels et~al.(2005)Michels, Saxena, and Ng]{michels2005high}
Jeff Michels, Ashutosh Saxena, and Andrew~Y Ng.
\newblock High speed obstacle avoidance using monocular vision and reinforcement learning.
\newblock In \emph{Proceedings of the 22nd international conference on Machine learning}, pages 593--600, 2005.

\bibitem[Mildenhall et~al.(2020)Mildenhall, Srinivasan, Tancik, Barron, Ramamoorthi, and Ng]{nerf}
Ben Mildenhall, Pratul~P. Srinivasan, Matthew Tancik, Jonathan~T. Barron, Ravi Ramamoorthi, and Ren Ng.
\newblock Nerf: Representing scenes as neural radiance fields for view synthesis.
\newblock In \emph{ECCV}, 2020.

\bibitem[Nagai et~al.(2002)Nagai, Naruse, Ikehara, and Kurematsu]{nagai2002hmm}
Takaaki Nagai, Takumi Naruse, Masaaki Ikehara, and Akira Kurematsu.
\newblock Hmm-based surface reconstruction from single images.
\newblock In \emph{Proceedings. International Conference on Image Processing}, pages II--II. IEEE, 2002.

\bibitem[Ning et~al.(2023)Ning, Li, Zhang, Wang, Geng, Dai, He, and Hu]{ait}
Jia Ning, Chen Li, Zheng Zhang, Chunyu Wang, Zigang Geng, Qi Dai, Kun He, and Han Hu.
\newblock All in tokens: Unifying output space of visual tasks via soft token.
\newblock In \emph{ICCV}, 2023.

\bibitem[Oquab et~al.(2023)Oquab, Darcet, Moutakanni, Vo, Szafraniec, Khalidov, Fernandez, Haziza, Massa, El-Nouby, et~al.]{dinov2}
Maxime Oquab, Timoth{\'e}e Darcet, Th{\'e}o Moutakanni, Huy Vo, Marc Szafraniec, Vasil Khalidov, Pierre Fernandez, Daniel Haziza, Francisco Massa, Alaaeldin El-Nouby, et~al.
\newblock Dinov2: Learning robust visual features without supervision.
\newblock \emph{TMLR}, 2023.

\bibitem[Patil et~al.(2022)Patil, Sakaridis, Liniger, and Van~Gool]{p3depth}
Vaishakh Patil, Christos Sakaridis, Alexander Liniger, and Luc Van~Gool.
\newblock P3depth: Monocular depth estimation with a piecewise planarity prior.
\newblock In \emph{CVPR}, 2022.

\bibitem[Ranftl et~al.(2021)Ranftl, Bochkovskiy, and Koltun]{dpt}
Ren{\'e} Ranftl, Alexey Bochkovskiy, and Vladlen Koltun.
\newblock Vision transformers for dense prediction.
\newblock In \emph{ICCV}, 2021.

\bibitem[Ranftl et~al.(2022)Ranftl, Lasinger, Hafner, Schindler, and Koltun]{midas}
Ren{\'e} Ranftl, Katrin Lasinger, David Hafner, Konrad Schindler, and Vladlen Koltun.
\newblock Towards robust monocular depth estimation: Mixing datasets for zero-shot cross-dataset transfer.
\newblock \emph{TPAMI}, 2022.

\bibitem[Roberts et~al.(2021)Roberts, Ramapuram, Ranjan, Kumar, Bautista, Paczan, Webb, and Susskind]{hypersim}
Mike Roberts, Jason Ramapuram, Anurag Ranjan, Atulit Kumar, Miguel~Angel Bautista, Nathan Paczan, Russ Webb, and Joshua~M Susskind.
\newblock Hypersim: A photorealistic synthetic dataset for holistic indoor scene understanding.
\newblock In \emph{ICCV}, 2021.

\bibitem[Rombach et~al.(2022)Rombach, Blattmann, Lorenz, Esser, and Ommer]{sd}
Robin Rombach, Andreas Blattmann, Dominik Lorenz, Patrick Esser, and Bj{\"o}rn Ommer.
\newblock High-resolution image synthesis with latent diffusion models.
\newblock In \emph{CVPR}, 2022.

\bibitem[Ros et~al.(2016)Ros, Sellart, Materzynska, Vazquez, and Lopez]{synthia}
German Ros, Laura Sellart, Joanna Materzynska, David Vazquez, and Antonio~M Lopez.
\newblock The synthia dataset: A large collection of synthetic images for semantic segmentation of urban scenes.
\newblock In \emph{Proceedings of the IEEE conference on computer vision and pattern recognition}, pages 3234--3243, 2016.

\bibitem[Saxena et~al.(2005)Saxena, Chung, and Ng]{saxena2005learning}
Ashutosh Saxena, Sung Chung, and Andrew Ng.
\newblock Learning depth from single monocular images.
\newblock \emph{Advances in neural information processing systems}, 18, 2005.

\bibitem[Shao et~al.(2023{\natexlab{a}})Shao, Pei, Chen, Wu, and Li]{nddepth}
Shuwei Shao, Zhongcai Pei, Weihai Chen, Xingming Wu, and Zhengguo Li.
\newblock Nddepth: Normal-distance assisted monocular depth estimation.
\newblock In \emph{ICCV}, 2023{\natexlab{a}}.

\bibitem[Shao et~al.(2023{\natexlab{b}})Shao, Pei, Wu, Liu, Chen, and Li]{iebins}
Shuwei Shao, Zhongcai Pei, Xingming Wu, Zhong Liu, Weihai Chen, and Zhengguo Li.
\newblock Iebins: Iterative elastic bins for monocular depth estimation.
\newblock In \emph{NeurIPS}, 2023{\natexlab{b}}.

\bibitem[Silberman et~al.(2012)Silberman, Hoiem, Kohli, and Fergus]{nyud}
Nathan Silberman, Derek Hoiem, Pushmeet Kohli, and Rob Fergus.
\newblock Indoor segmentation and support inference from rgbd images.
\newblock In \emph{ECCV}, 2012.

\bibitem[Song et~al.(2015)Song, Lichtenberg, and Xiao]{sunrgbd}
Shuran Song, Samuel~P Lichtenberg, and Jianxiong Xiao.
\newblock Sun rgb-d: A rgb-d scene understanding benchmark suite.
\newblock In \emph{CVPR}, 2015.

\bibitem[Vasiljevic et~al.(2019)Vasiljevic, Kolkin, Zhang, Luo, Wang, Dai, Daniele, Mostajabi, Basart, Walter, et~al.]{diode}
Igor Vasiljevic, Nick Kolkin, Shanyi Zhang, Ruotian Luo, Haochen Wang, Falcon~Z Dai, Andrea~F Daniele, Mohammadreza Mostajabi, Steven Basart, Matthew~R Walter, et~al.
\newblock Diode: A dense indoor and outdoor depth dataset.
\newblock \emph{arXiv:1908.00463}, 2019.

\bibitem[Wang et~al.(2025)Wang, Zhang, Chen, Jun, and Liu]{pddepth}
Huanan Wang, Xinyu Zhang, Zhengxian Chen, Li Jun, and Huaping Liu.
\newblock Pddepth: Pose decoupled monocular depth estimation for roadside perception system.
\newblock \emph{IEEE Transactions on Circuits and Systems for Video Technology}, 2025.

\bibitem[Yang et~al.(2024{\natexlab{a}})Yang, Huang, Yin, Shen, Liu, He, Lin, Ouyang, and He]{yang2024depthvideoscalablesynthetic}
Honghui Yang, Di Huang, Wei Yin, Chunhua Shen, Haifeng Liu, Xiaofei He, Binbin Lin, Wanli Ouyang, and Tong He.
\newblock Depth any video with scalable synthetic data, 2024{\natexlab{a}}.

\bibitem[Yang et~al.(2024{\natexlab{b}})Yang, Kang, Huang, Xu, Feng, and Zhao]{depth_anything}
Lihe Yang, Bingyi Kang, Zilong Huang, Xiaogang Xu, Jiashi Feng, and Hengshuang Zhao.
\newblock Depth anything: Unleashing the power of large-scale unlabeled data.
\newblock In \emph{CVPR}, 2024{\natexlab{b}}.

\bibitem[Yang et~al.(2024{\natexlab{c}})Yang, Kang, Huang, Xu, Feng, and Zhao]{depthanythingv1}
Lihe Yang, Bingyi Kang, Zilong Huang, Xiaogang Xu, Jiashi Feng, and Hengshuang Zhao.
\newblock Depth anything: Unleashing the power of large-scale unlabeled data.
\newblock In \emph{Proceedings of the IEEE/CVF Conference on Computer Vision and Pattern Recognition}, pages 10371--10381, 2024{\natexlab{c}}.

\bibitem[Yang et~al.(2025)Yang, Kang, Huang, Zhao, Xu, Feng, and Zhao]{depthanythingv2}
Lihe Yang, Bingyi Kang, Zilong Huang, Zhen Zhao, Xiaogang Xu, Jiashi Feng, and Hengshuang Zhao.
\newblock Depth anything v2.
\newblock \emph{Advances in Neural Information Processing Systems}, 37:\penalty0 21875--21911, 2025.

\bibitem[Yao et~al.(2020)Yao, Luo, Li, Zhang, Ren, Zhou, Fang, and Quan]{blendedmvs}
Yao Yao, Zixin Luo, Shiwei Li, Jingyang Zhang, Yufan Ren, Lei Zhou, Tian Fang, and Long Quan.
\newblock Blendedmvs: A large-scale dataset for generalized multi-view stereo networks.
\newblock In \emph{CVPR}, 2020.

\bibitem[Yin et~al.(2023)Yin, Zhang, Chen, Cai, Yu, Wang, Chen, and Shen]{metric3d}
Wei Yin, Chi Zhang, Hao Chen, Zhipeng Cai, Gang Yu, Kaixuan Wang, Xiaozhi Chen, and Chunhua Shen.
\newblock Metric3d: Towards zero-shot metric 3d prediction from a single image.
\newblock In \emph{ICCV}, 2023.

\bibitem[Yuan et~al.(2022)Yuan, Gu, Dai, Zhu, and Tan]{newcrfs}
Weihao Yuan, Xiaodong Gu, Zuozhuo Dai, Siyu Zhu, and Ping Tan.
\newblock New crfs: Neural window fully-connected crfs for monocular depth estimation.
\newblock \emph{arXiv:2203.01502}, 2022.

\bibitem[Zhang et~al.(2018)Zhang, Cui, Xu, Jie, Li, and Yang]{zhang2018joint}
Zhenyu Zhang, Zhen Cui, Chunyan Xu, Zequn Jie, Xiang Li, and Jian Yang.
\newblock Joint task-recursive learning for semantic segmentation and depth estimation.
\newblock In \emph{Proceedings of the European conference on computer vision (ECCV)}, pages 235--251, 2018.

\bibitem[Zhao et~al.(2023)Zhao, Rao, Liu, Liu, Zhou, and Lu]{vpd}
Wenliang Zhao, Yongming Rao, Zuyan Liu, Benlin Liu, Jie Zhou, and Jiwen Lu.
\newblock Unleashing text-to-image diffusion models for visual perception.
\newblock In \emph{ICCV}, 2023.

\bibitem[Zhu et~al.(2024)Zhu, Wang, Song, Liu, Zhang, and Zhang]{scaledepth}
Ruijie Zhu, Chuxin Wang, Ziyang Song, Li Liu, Tianzhu Zhang, and Yongdong Zhang.
\newblock Scaledepth: Decomposing metric depth estimation into scale prediction and relative depth estimation.
\newblock \emph{arXiv preprint arXiv:2407.08187}, 2024.

\end{thebibliography}
}

\clearpage
\setcounter{page}{1}
\maketitlesupplementary

\section{Datasets} \label{sec:Datasets}
We train our model on a diverse set of both real and synthetic datasets that span a variety of ranges and scenes, as listed in Table~\ref{dataset}. By leveraging this extensive training data, our model is able to effectively capture the complexities of different environments. As a result, we achieve depth map estimations that not only have a high dynamic range but also exhibit sharp, well-defined edges. This allows for more accurate and reliable depth perception across various contexts, enhancing the overall quality of 3D reconstruction and scene understanding.

\begin{table}[!ht]
    \centering
    \begin{tabular}{cccc}
    \toprule
        Datasets & Scenes & Depth From & Pair Size \\ \hline
        ETH3D~ & outdoor & real & 454 \\ 
        DSEC & outdoor & real & 63931 \\ 
        vkitti & outdoor & synthetic & 20000 \\ 
        MVS-Synth & outdoor & synthetic & 12000 \\ 
        hypersim & indoor & synthetic & 142350 \\
        DIML & out/indoor & real & 70030 \\ 
        cityscapes & outdoor & real & 174998 \\ 
        
    \bottomrule
    \end{tabular}
    \caption{
    \textbf{Datasets for training.}
    We train our model on real and synthetic datasets across different ranges and scenes, achieving high dynamic range and sharp-edged depth map estimation.}
    \label{dataset}
\end{table}

\section{More Visualization Results}
\begin{figure*}[t]
  \centering
  \includegraphics[width=1.0\linewidth]{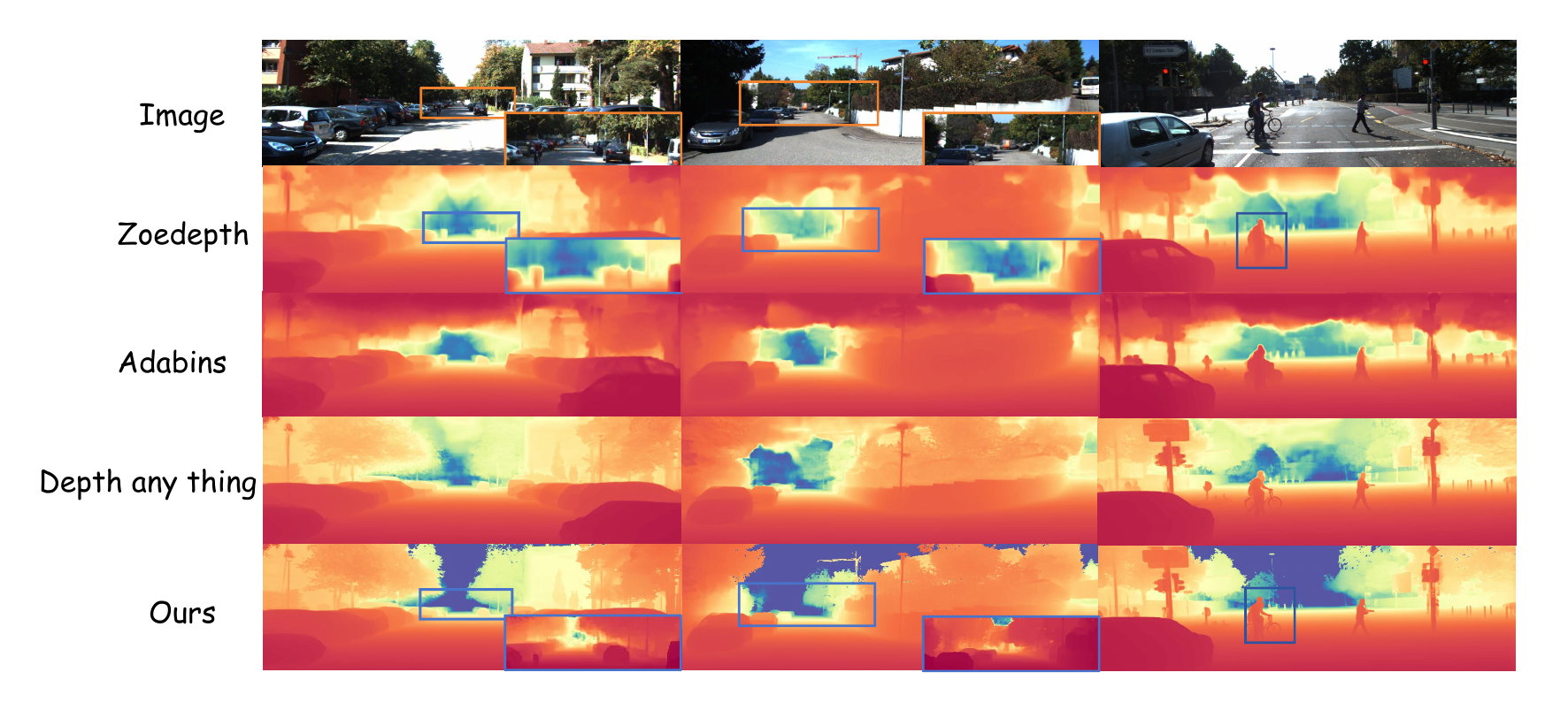}
   \caption{\textbf{Qualitative comparisons of depth predictions on the outdoor dataset KITTI}.
   It can be observed that our method performs better in predicting details both in the near range and far distance. To improve visualization clarity, we cropped the part beyond 80 meters, but we still achieve good predictions for that portion.}
   \label{fig:comp_depth_pred_outdoor}
   
\end{figure*}

\begin{figure*}[t]
  \centering
  \includegraphics[width=1.0\linewidth]{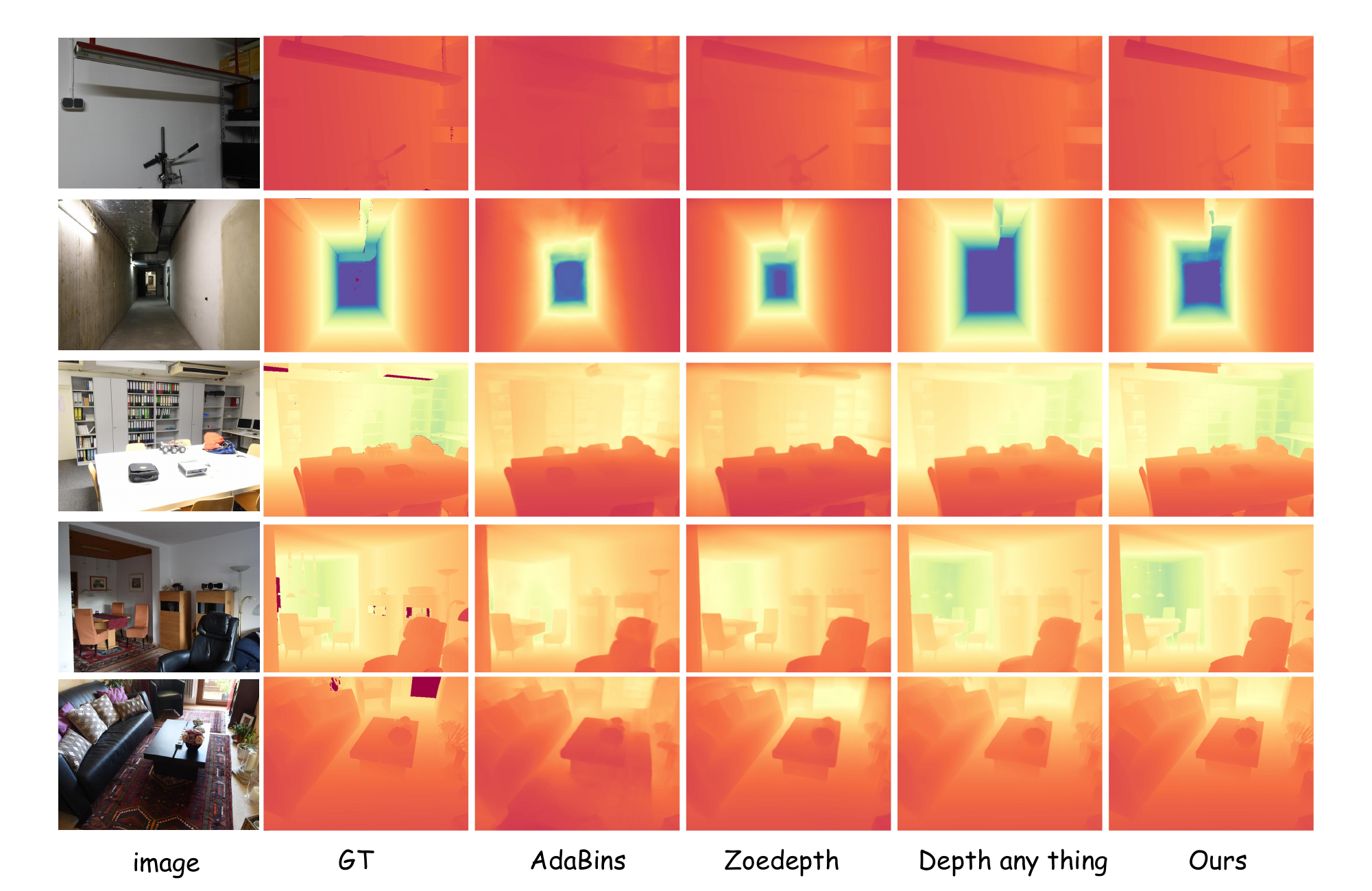}
   \caption{\textbf{Qualitative comparisons of depth predictions on the indoor dataset iBims}. 
   It can be observed that our method performs better in predicting details in the near range (e.g., the first row). Moreover, our predictions for long distances in indoor scenes are more accurate (e.g., the second row).}
   \label{fig:comp_ibims}
\end{figure*}

\begin{figure*}[t]
  \centering
  \includegraphics[width=1.0\linewidth]{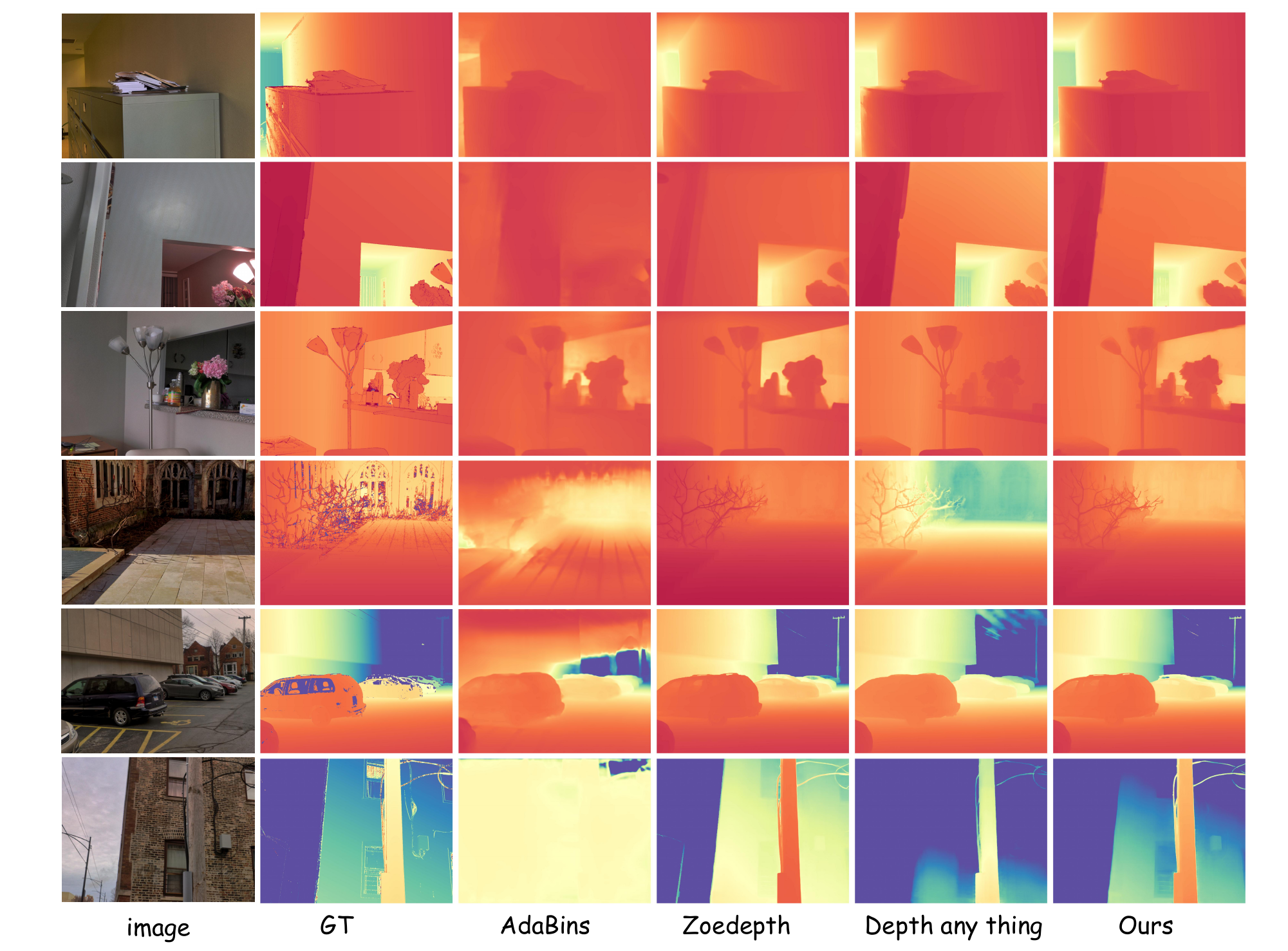}
   \caption{\textbf{Qualitative comparisons on diode}. Our test results on the DIODE dataset, which is a real-world dataset containing both indoor and outdoor scenes, demonstrate that our method achieves high accuracy in mixed scenes.}
   \label{fig:comp_diode}
\end{figure*}

We further present visual comparisons on both indoor and outdoor datasets in Fig.\ref{fig:comp_depth_pred_outdoor}, Fig.\ref{fig:comp_ibims}, and Fig. \ref{fig:comp_diode}. 
Our method not only maintains high depth accuracy but also preserves fine structural details, resulting in high-quality scene predictions. Moreover, it generalizes well across diverse environments using a single model, demonstrating strong adaptability and robustness. This combination of depth fidelity, detail preservation, and cross-scene generalization distinguishes our approach in terms of both efficiency and effectiveness.


\begin{figure*}[t]
  \centering
  \begin{overpic}[width=0.9\linewidth]{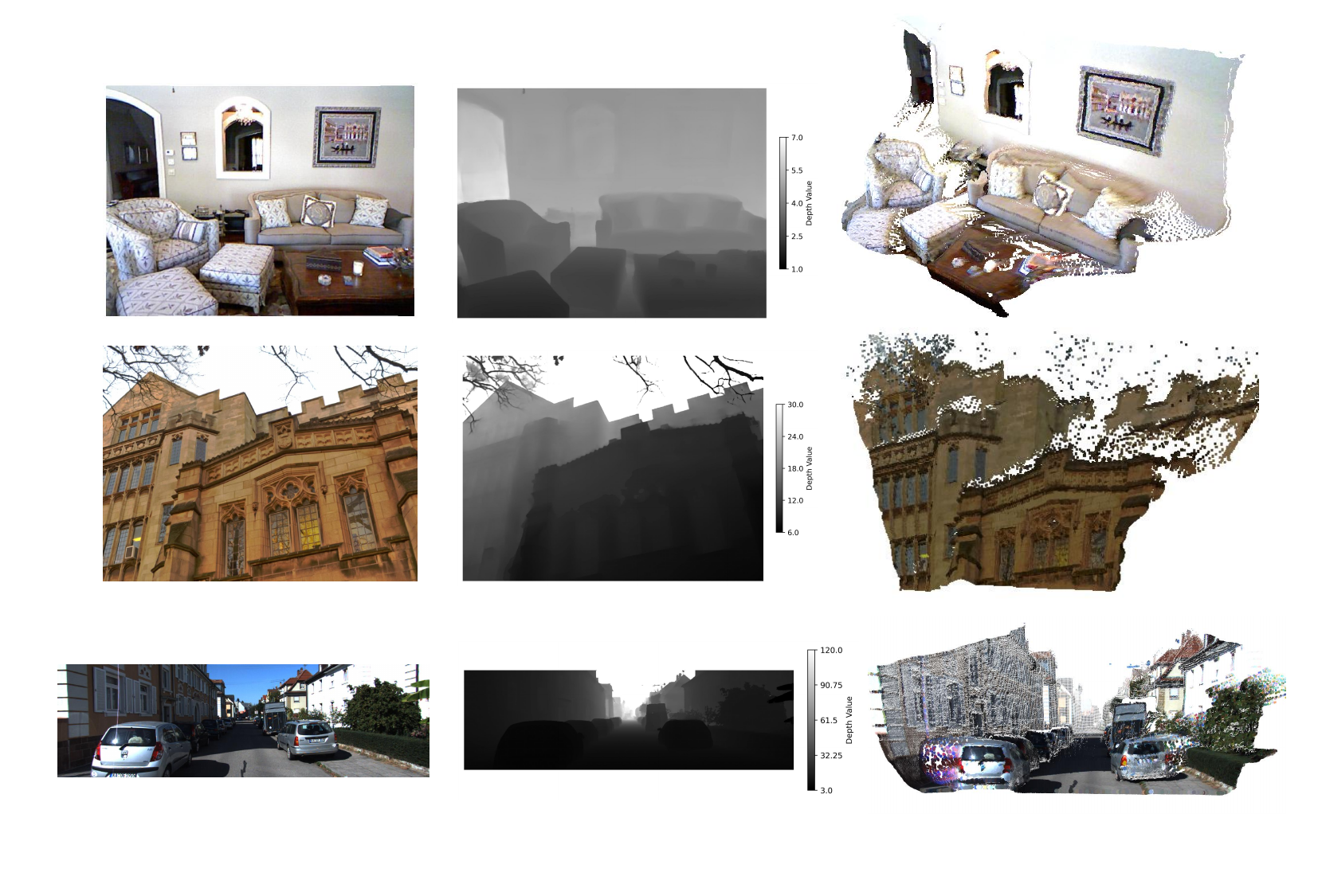}
  \end{overpic}
  \vspace{-18pt}
   \caption{\textbf{3D Reconstruction based on Depth Estimation} We obtain the metric depth of different scenes through depth estimation and acquire the point cloud information of the scene via back-projection. This allows us not only to obtain an accurate scale of the scene but also to achieve absolute depth estimation across different scenes.
   }
   \label{fig:app_recon}
\end{figure*}

\section{Evaluation Metrics} 

We evaluate depth prediction in metric depth space \(\mathbf{d}\) using several standard metrics. Specifically, we compute:

Absolute Relative Error (REL):
  \[
  \text{REL} = \frac{1}{M}\sum_{i=1}^{M} \frac{|\mathbf{d}_{i} - \hat{\mathbf{d}}_{i}|}{\mathbf{d}_{i}}
  \]

Root Mean Squared Error (RMSE):
  \[
  \text{RMSE} = \left[\frac{1}{M}\sum_{i=1}^{M} |\mathbf{d}_{i} - \hat{\mathbf{d}}_{i}|^2\right]^{\frac{1}{2}}
  \]

Average \(\log_{10}\) Error:
  \[
  \log_{10} = \frac{1}{M} \sum_{i=1}^{M} \left| \log_{10}{\mathbf{d}_{i}} - \log_{10}{\hat{\mathbf{d}}_{i}} \right|
  \]

Threshold Accuracy (\(\delta_n\)):
  The percentage of pixels such that  
  \[
  \max\left(\frac{\mathbf{d}_{i}}{\hat{\mathbf{d}}_{i}}, \frac{\hat{\mathbf{d}}_{i}}{\mathbf{d}_{i}}\right) < 1.25^n \quad \text{for } n = 1, 2, 3
  \]

Here, \(\mathbf{d}_i\) and \(\hat{\mathbf{d}}_i\) denote the ground-truth and predicted depths at pixel \(i\), respectively, and \(M\) is the total number of valid pixels in the image.

We cap the evaluation depth at 10 meters for indoor datasets and 80 meters for outdoor datasets. Final predictions are obtained by averaging the depth map of the original image with that of its horizontally flipped counterpart, and evaluations are performed at the original ground-truth resolution.

\section{Application - Monocular Reconstruction}
In monocular surface reconstruction, we use our predicted metric depth maps to directly reproject the 2D depth information into 3D space. By utilizing the camera intrinsics, we can convert the depth values to real-world coordinates. This allows us to construct a 3D point cloud of the scene from a single image. Our approach’s ability to provide accurate and unified depth predictions across varying scene scales makes it useful for reconstructing detailed 3D structures from monocular images, even in scenes with significant depth variations, such as indoor and outdoor environments. Please refer to Fig.~\ref{app_recon} in the supplemental for more details.
\end{document}